\documentclass[10pt,twocolumn,letterpaper]{article}

\usepackage[pagenumbers]{cvpr} 
\usepackage{indentfirst}
\usepackage{multirow}
\usepackage[dvipsnames]{xcolor}

\definecolor{cvprblue}{rgb}{0.21,0.49,0.74}
\usepackage[pagebackref,breaklinks,colorlinks,citecolor=cvprblue]{hyperref}

\def\paperID{14508} 
\def\confName{CVPR}
\def\confYear{2024}

\title{Arbitrary-Scale Image Generation and Upsampling using Latent Diffusion Model and Implicit Neural Decoder}

\author{Jinseok Kim$^{1,3}$\qquad Tae-Kyun Kim$^{1,2}$\\
$^1$KAIST\qquad $^2$Imperial College London\\
$^3$AI Lab, LG Electronics\\
}

\begin{document}
\twocolumn[{%
\renewcommand\twocolumn[1][]{#1}%
\maketitle
\begin{center}
\vspace{-1.3em}
    \centering
    \captionsetup{type=figure}
    \includegraphics[width=\textwidth]{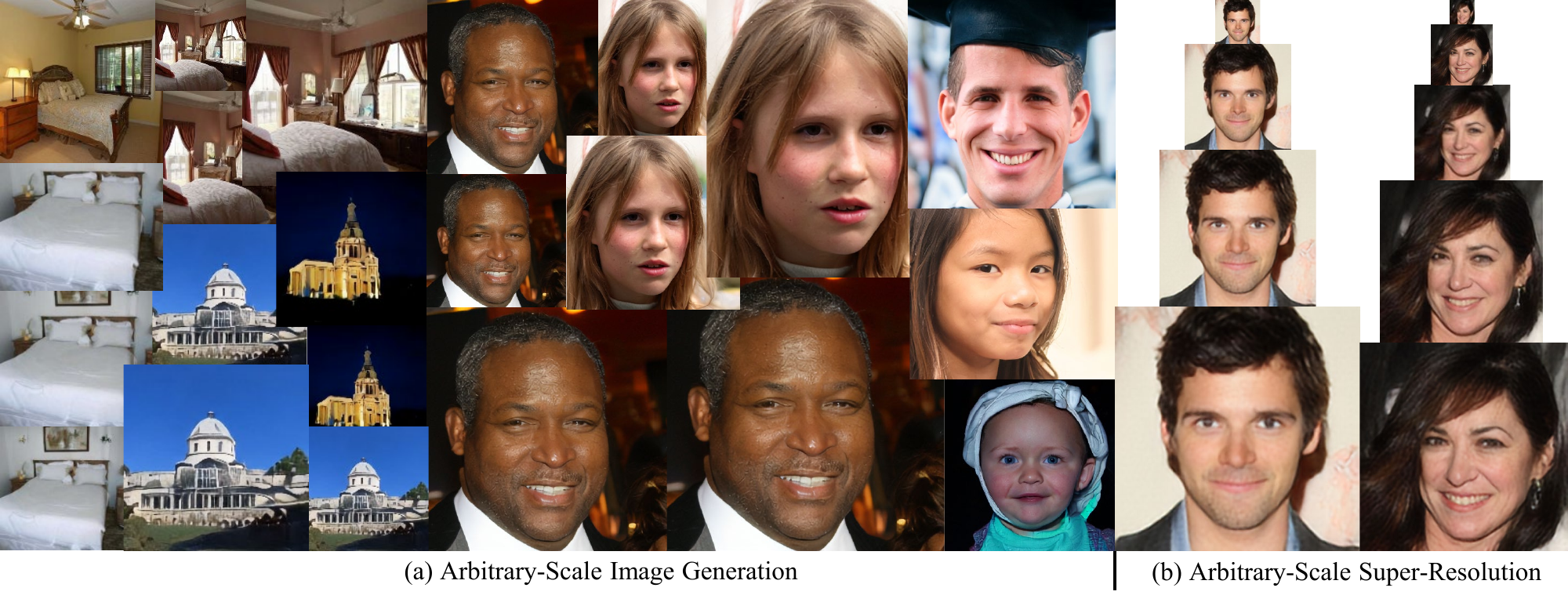}
    \vspace{-1.3em}
    \captionof{figure}{The proposed method generates novel images and 
    super-resolves low-resolution images at arbitrary-scales with high fidelity, diversity, and fast inference speed.
    }
    \label{fig:teaser}
\end{center}%
}]

\begin{abstract}
\vspace{-1.25em}
Super-resolution (SR) and image generation are important tasks in computer vision and are widely adopted in real-world applications.
Most existing methods, however, generate images only at fixed-scale magnification and suffer from over-smoothing and artifacts.
Additionally, they do not offer enough diversity of output images nor image consistency at different scales.
Most relevant work applied Implicit Neural Representation (INR) to the denoising diffusion model to obtain continuous-resolution yet diverse and high-quality SR results.
Since this model operates in the image space, the larger the resolution of image is produced, the more memory and inference time is required, and it also does not maintain scale-specific consistency.
We propose a novel pipeline that can super-resolve an input image or generate from a random noise a novel image at arbitrary scales. The method consists of a pre-trained auto-encoder, a latent diffusion model, and an implicit neural decoder, and their learning strategies.
The proposed method adopts diffusion processes in a latent space, thus efficient, yet aligned with output image space decoded by MLPs at arbitrary scales. 
More specifically, our arbitrary-scale decoder is designed by the symmetric decoder w/o up-scaling from the pre-trained auto-encoder, and  Local Implicit Image Function (LIIF) in series. The latent diffusion process is learnt by the denoising and the alignment losses jointly. Errors in output images are backpropagated via the fixed decoder, improving the quality of output images.
In the extensive experiments using multiple public benchmarks on the two tasks i.e. image super-resolution and novel image generation at arbitrary scales, the proposed method outperforms relevant methods in metrics of image quality, diversity and scale consistency. It is significantly better than the relevant prior-art in the inference speed and memory usage. 
\end{abstract}
\vspace{-1.24em}
    
\section{Introduction}
\label{sec:intro}

Super-resolution (SR) is the task of restoring a high-resolution (HR) image by estimating the high-frequency details of an input low-resolution (LR) image.
SR has applications in various fields, including medical imaging, satellite imaging, surveillance, and digital photography.
It helps to enhance the visual quality of images, making them more suitable for analysis, interpretation, or presentation.
SR is a long-standing study in the field of computer vision; it is still a challenging problem.
Since multiple HR images can be converted to the same LR image, obtaining the original HR image given a LR image is an `ill-posed problem' in which there is no single correct answer. Therefore, the SR models should be able to generate diverse HR images with high perceptual quality while maintaining the coarse feature of the LR image well.

Image generation is the task of generating new images from an existing dataset.
It has a wide range of applications, including data augmentation for machine learning, computer graphics, art creation, content creation for virtual environments, and more.
However, high-dimensional data that look more realistic and contain fine details are required, and diverse images must be generated while maintaining high quality.

\begin{figure}[t]
  \centering
  \begin{subfigure}{1.0\linewidth}
     \includegraphics[width=1.0\linewidth]{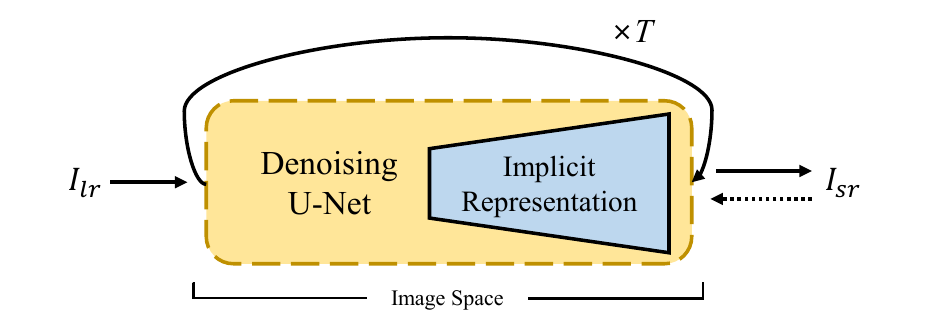}
     \caption{IDM}
     \label{fig:architecture_comparison_idm}
  \end{subfigure}
  \vfill
  \begin{subfigure}{1.0\linewidth}
     \includegraphics[width=1.0\linewidth]{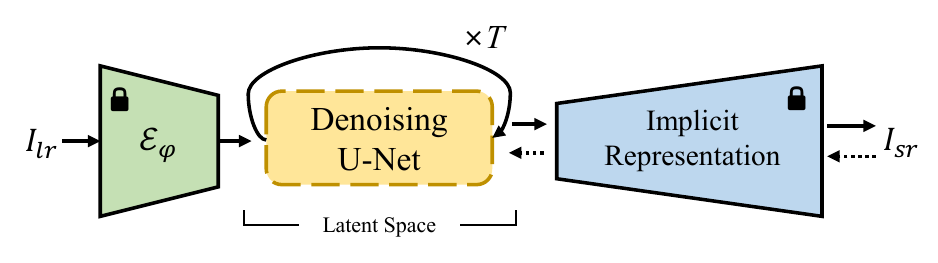}
     \caption{Ours}
     \label{fig:architecture_comparison_ours}
  \end{subfigure}
  \caption{Model structure comparison with IDM. The solid arrow represents the inference process, and the dotted arrow represents the error backpropagation process.}
  \label{fig:architecture_comparison}
\end{figure}

For both super-resolution and image generation tasks, several methodologies using deep neural networks have been proposed.
The emergence of GAN and Diffusion Models (DMs) has brought a new paradigm to these areas.
Various methods~\cite{MSG-GAN, ProGAN, CDM} based on them have enabled high-quality image synthesis at various scales.
However, these models only work at a fixed integer scale factor(2$\times$, 4$\times$, 8$\times$) and do not address consistency of results across different scales.
As SR methods, regression-based models, such as EDSR~\cite{EDSR}, RRDB~\cite{RRDB}, RCAN~\cite{RCAN} and SwinIR~\cite{SwinIR}, learn a mapping from LR images to HR images in an end-to-end manner.
Unfortunately, these models also work solely on specific trained scales. Recent methods have been developed for upscaling on continuous scales, using an implicit image function. Their MLP-based decoder takes an arbitrary query point in 2D pixel space and yields predicted pixel colors, such as Meta-SR~\cite{Meta-SR}, LTE~\cite{LTE}, and LIIF~\cite{LIIF}. 
However, since they often suffer from duller edges and over-smoothing details, the perceptual quality is not as high as those of fixed-scale methods. 
The most relevant work to this study is IDM~\cite{IDM}.
It applies Implicit Neural Representation (INR) to the denoising diffusion model to obtain continuous-resolution yet diverse and high-quality SR results.
This model, however, applies implicit representation to the denoising U-Net at each diffusion step, raising the complexity of training/inference and the need of a large memory.
See \cref{fig:architecture_comparison_idm}.
The method also yields poor scale consistency.
The aforementioned methods are limited to SR than image generation.
Continuous scale image generation has also received an attention.
Ntavelis \etal~\cite{ScaleParty} proposed a scale-consistent positional encoding to generate images of arbitrary-scales with high fidelity based on GANs.
In contrast, we adopt a diffusion probabilistic model to enable diverse yet improved output qualities. Different from IDM~\cite{IDM}, our neural decoder is taken out of the diffusion process and the diffusion is done in a latent space(\cref{fig:architecture_comparison_ours}).
See also Section \ref{sec:related} for the related works. 

To address the aforementioned issues, this paper proposes a diffusion-based model for arbitrary-scale image upsampling and image synthesis simultaneously.
The main contributions of this paper are as follows:
\begin{itemize}
    \item We design a decoder that combines an auto-decoder and MLPs to generate images of arbitrary scales from the latent space.
    \item We introduce a two-stage model alignment process to reduce errors and misalignment between a Latent Diffusion Model (LDM) and image decoder that may occur during training.
    \item The proposed method enables faster and more efficient image generation compared to the other diffusion-based super-resolution model, as well as offers high fidelity and diverse output images. 
\end{itemize}

\section{Related Work}
\label{sec:related}

\textbf{Diffusion Models.}
The diffusion model is one of the most powerful and widely used generative models.
It eliminates noise from a noisy input (which usually follows a Gaussian distribution) and transforms it into sample data that follow the desired data distribution.
These operations are often performed directly in the pixel space.
However, recent developments have been made called Latent Diffusion Models (LDMs)~\cite{LDM}, where LDMs perform diffusion operations in a lower-dimensional latent space, being computationally efficient and capturing complex patterns more effectively.
Diffusion models enable the production of results with high quality and diversity due to its stochastic nature.
Recently, diffusion-based models have been extended to various domains such as 3D data, demonstrations, and voices, and have achieved state-of-the-art results on several benchmarks~\cite{iEdit, InterHandGen, PromptAug, LLDiffusion}.

\textbf{Arbitrary-Scale Image Generation.}
Anokhin \etal~\cite{CIPS} and Skorokhodov \etal~\cite{INR-GAN} introduced CIPS and INR-GAN respectively, in which  MLPs were applied instead of using spatial convolutions commonly used in existing GAN-based models.
They allow models learned on single scale to produce images on several different scales.
Ntavelis \etal~\cite{ScaleParty} proposed a scale-consistent positional encoding to generate images of arbitrary-scales with high fidelity and maintaining consistency according to the scale.

A standard diffusion model uses U-Net~\cite{U-Net} structure consisting of a convolutional network.
Therefore, as long as certain conditions are met (\eg~ resolution in multiples of 8), there are no restrictions on the input size.
Despite supporting flexible input sizes, they are trained with a fixed image size, which can lead to out-of-domain problems when generating images of different sizes.
For example, a diffusion model trained with 256$\times$256 face datasets can only generate the faces images of that size.
Recently, Wang \etal~\cite{DDNM_Plus} and Bar-Tal \etal~\cite{MultiDiffusion} introduced the model that can generate high-quality arbitrary-resolution images by generating multiple patch images and blending them.
However, they require multiple inferences to generate a single image and do not address the problem of generating the same image only at different scales, i.e. arbitrary-scale super-resolution. 

\textbf{Arbitrary-Scale Super-Resolution.}
Since the inception of MetaSR~\cite{Meta-SR}, there has been a surge in the exploration of various approaches for single-model arbitrary-scale super-resolution.
LIIF~\cite{LIIF} employed an implicit decoding function that takes a 2D coordinate and neighboring feature vectors as input and produces the RGB value of the corresponding pixel.
However, the perceptual quality was not as good as that of high PSNR due to the duller edges and over-smoothing details.
Additionally, these approaches do not provide enough diversity of output images or image consistency at different scales.

By conditioning the LR images on the diffusion models, they can be extended to super-resolution models.
CDM~\cite{CDM} and SR3~\cite{SR3} gradually upsampled LR images into HR images.
However, they require training multiple networks separately at each scale.
PDDPM~\cite{PDDPM} enabled upsampling at multiple scales with a single model by exploiting the positional embedding, but the aforementioned models still only worked at fixed integer scales.
Recently, IDM~\cite{IDM} applied INR to the U-Net decoder to allow generating images on arbitrary scales.
Although it has shown impressive performance, the larger the image is produced, the lower the inference speed, and the exponentially more memory is used, because it has to pass through MLPs repeatedly in each reverse process.
In this paper, we applied diffusion processes on a latent space then a MLP-based decoder super-resolves the denoised latent vector to an arbitrary-scale image. The diffusion model and decoder can be separately learnt, offering a much simpler architecture.

\section{Preliminary}
\label{sec:preliminary}

\begin{figure*}[t!]
  \centering
  \includegraphics[width=1.0\linewidth]{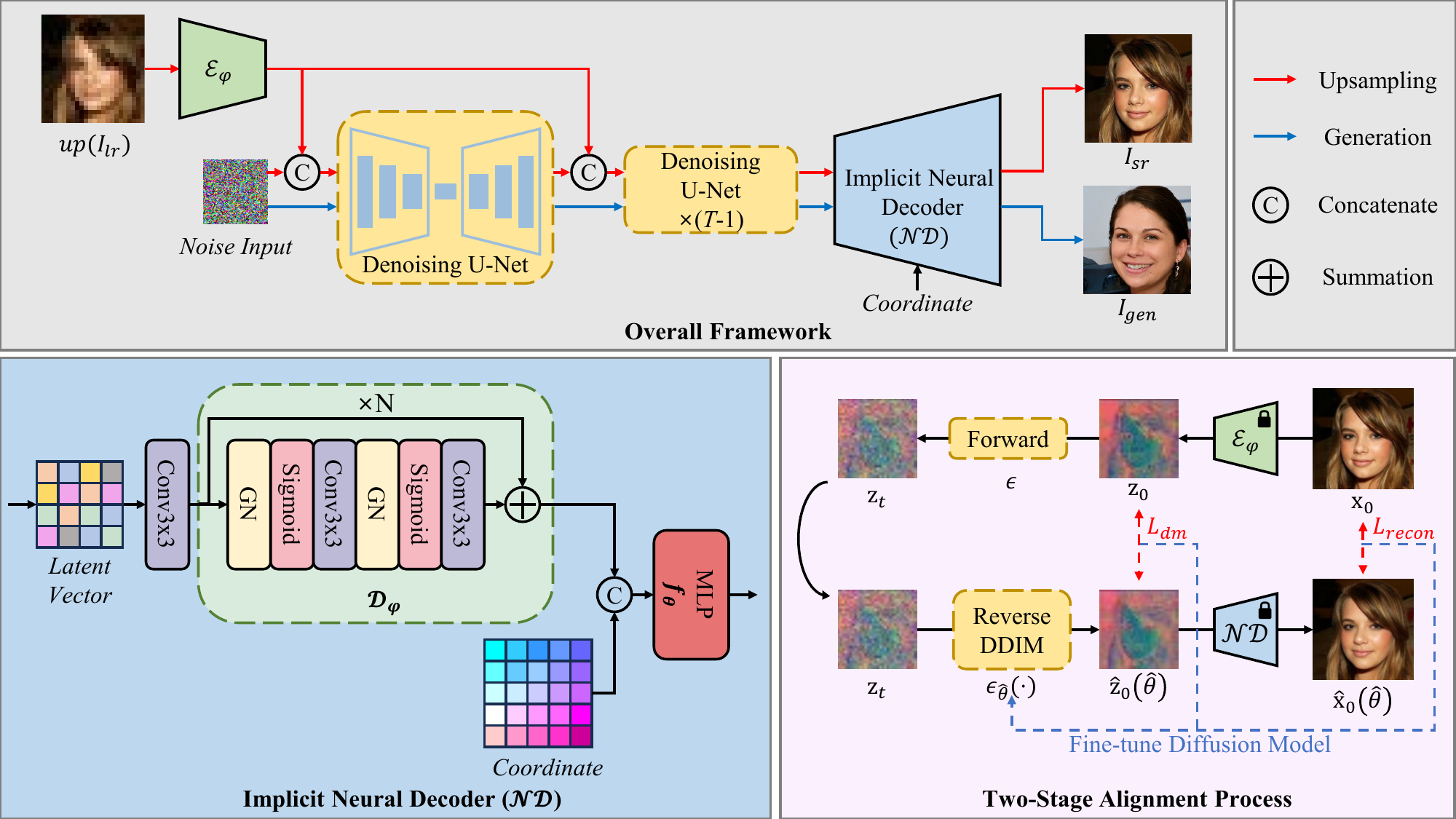}
  \vspace{-2.0em}
  \caption{
  \textbf{Upper Part:} Overall process of proposed networks.
  \textcolor{red}{Red line} is a super-resolution process, and \textcolor{blue}{Blue line} is a generation process.
  \textbf{Lower Left Part:} Detail architecture of Implicit Neural Decoder.
  It contains a series of auto-decoder $\mathcal{D}_{\varphi}$ and a neural decoding function $f_{\theta}$.
  \textbf{Lower Right Part:} Pipeline of two-stage alignment process.
  }
  \vspace{-1.5em}
  \label{fig:model_architecture}
\end{figure*}

\subsection{Latent Diffusion Models}
The core idea behind Latent Diffusion Models (LDMs) is to train a diffusion model in the latent space of a pre-trained auto-encoder, which allows for high-quality image synthesis with a flexible range of styles and resolutions. 
The diffusion process consists of a forward process and a reverse process.
The forward process gradually adds random noise into the data, whereas the reverse process constructs desired data samples from the noise.
The forward process is a fixed process and the noise latent z at time step $t$ can be expressed as:
\vspace{-0.5em}
\begin{equation}
\vspace{-0.5em}
\begin{split}
  \text{z}_t &\ = \sqrt{\bar{\alpha}_{t}} \text{z}_0 + \sqrt{1 - \bar{\alpha}_{t}} \epsilon, \quad \epsilon \sim \mathcal{N}(0, \text{I}), 
  \label{eq:ldm_forward}
\end{split}
\end{equation}
where, $\bar{\alpha}_{t} := \prod_{s=1}^{t}(1-\beta_{t})$ and $\left\{ \beta_{t} \right\}_{t=0}^{T}$ is a variance schedule. 
In contrast, the reverse process is the inverse of the forward process. 
However, the posterior distribution $p(\text{z}_{t-1}|\text{z}_{t})$ is intractable since it necessitates knowledge of the distribution encompassing all potential images to compute this conditional probability.
Therefore, the distribution is approximated through a neural network, and can be expressed using the Bayesian theorem as follows:
\vspace{-0.5em}
\begin{equation}
\vspace{-0.5em}
\begin{split}
   p_{\theta}(\text{z}_{t-1}|\text{z}_{t}) = \mathcal{N}(\text{z}_{t-1}; \mu_{t}(\text{z}_{t}, \text{z}_{0}), \sigma_{t}^{2}\text{I}),
  \label{eq:ldm_posterior}
\end{split}
\end{equation}
where
\vspace{-0.5em}
\begin{equation}
\vspace{-0.5em}
\begin{split}
   \mu_{t}(\text{z}_{t}, \text{z}_{0}) =\frac{1}{\sqrt{\alpha_{t}}} \left( \text{z}_{t}+\frac{\beta_{t}}{\sqrt{1 -\bar{\alpha}_{t}}} \epsilon \right), \quad \sigma_{t}^{2} = \beta_{t}.
  \label{eq:ldm_mean_variance}
\end{split}
\end{equation}
The authors of DDPM~\cite{DDPM} have shown that better results can be obtained by predicting noise at each time step t.
Therefore, the objective function of the diffusion model \( \epsilon_\theta\left( \cdot\right) \) can be expressed as:
\vspace{-0.5em}
\begin{equation}
\vspace{-0.5em}
\begin{split}
  L_\theta &\ = \left\| \epsilon - \epsilon_\theta\left( \text{z}_t, t \right) \right\|_{2}^{2}
  \label{eq:ldm_loss}
\end{split}
\end{equation}
\subsection{Local Implicit Image Function}
Local Implicit Image Function (LIIF)~\cite{LIIF} is a technique for representing images in a continuous way for arbitrary scale image upsampling.
It uses a decoding function that takes a 2D coordinate and neighboring feature vectors as input and outputs the RGB value of the corresponding pixel.
In LIIF representation, the RGB values of continuous images \(I\) at coordinates \(c \) are defined as
\begin{equation}
\begin{split}
  I(c) &\ = f_{\theta}(\text{z}^{\ast},c^{\ast}), 
  \label{eq:liif}
\end{split}
\end{equation}
where \(\text{z}^{*} \) is the interpolated feature vector obtained by calculating the nearest Euclidean distance from \(\text{z} \) and through the relative coordinate $c^{\ast}$ to image domain. $f_{\theta}$ is a decoding function parameterized as a MLP, and shared by all images.

It also improved the quality of continuous representation through other techniques such as feature unfolding, local ensemble and cell decoding.
In conclusion, LIIF leaned continuous representation by self-supervised learning and it can naturally exploit the information provided in ground-truths in different resolutions, even extrapolate to $\times$30 higher resolution.

\section{Method}
\vspace{-0.7em}
\label{sec:method}
\subsection{Overview}
\vspace{-0.4em}
We propose a simple architecture that combines the LDM and LIIF decoder for both arbitrary-scale SR and image generation (Fig. \ref{fig:model_architecture}).
As the encoder/decoder is fixed and the diffusion process is independently applied to the latent space in the LDM (note this significantly decreases learning complexity yet LDMs have shown outperforming other variants of DMs), we follow this stage learning strategy than end-to-end learning. An auto-encoder consisting of an encoder and a symmetric decoder w/o upsampling is pre-trained.
Our implicit neural decoder combines the convolutional decoder from the auto-encoder and MLP-based decoder, that can map to arbitrary-scale output images.
The diffusion process is done in a latent space, freezing the encoder and the neural decoder.
The pre-trained decoder successfully super-resolves the denoised latent vector to any scale output images.
In order to align the image space and latent space better, we backpropagate the image losses via the decoder to the $0$-th diffusion step and any $t$-th step ~\cite{LDM,PASD,ProgressiveDistillation,DiffusionCLIP}.
Thus, the latent diffusion occurs with its original denoising objective plus the image loss.
We experimented diverse variants of the proposed architecture, including the end-to-end learning, different decoder architectures, combinatorial loss functions (see the Supplementary material), and the proposed pipeline delivers the best results on all tasks and benchmarks.
The simple architecture benefits in terms of efficient inference to arbitrary scales and learning complexity as well as image quality and diversity.

\subsection{Encoder-Decoder}
\vspace{-0.4em}
Our model is composed of the encoder part, denoising diffusion part, decoder part.
The encoder-decoder structure follows a basic auto-encoder with convolutional and transposed convolutional neural networks, the encoder $\mathcal{E}_\varphi$ extracts the image \(\text{x} \in \mathbb{R}^{H \times W \times 3} \) into the latent vector \(\text{z} \in \mathbb{R}^{h \times w \times c} \), and the decoder part reconstructs the \(\text{z} \) back into the image space.
To improve the decoding ability, 
the symmetric structure decoder in the auto-encoder was used (this exploits $N$ ResBlocks where GN and Conv layers repeat), and only the upsamping layer was removed. Additionally, a MLP is combined behind it to generate arbitrary-scale images (Fig. \ref{fig:model_architecture}).

In other words, the auto-decoder is learned to increase the amount of meaningful information by expanding the dimension of the latent feature vector, and the MLP is learned to map the latent to RGB values along with output coordinates.
The feature vectors sequentially pass through the symmetric convolutional decoder and MLP and are decoded into an image.
Finally, to represent the RGB values of continuous images \(I\), we propose to modify LIIF (on \cref{eq:liif}) as
\vspace{-0.5em}
\begin{equation}
\vspace{-0.5em}
\begin{split}
  I(c) = \mathcal{ND}(\text{z}, c^{\ast}) = f_\theta(\mathcal{D}_{\varphi}(\text{z}), c^{\ast}).
  \label{eq:MLP_w_decoder}
\end{split}
\end{equation}
$\mathcal{D}_{\varphi}$ is the decoder network from the auto-encoder, $\text{z}$ is a sampled latent vector by the diffusion model and $c^*$ the relative pixel coordinate.
The latent vector $\text{z}$ is primarily decoded by the decoder $\mathcal{D}_{\varphi}$, and then interpolated through the Euclidean distance from $c^*$.
$f_\theta$ is modeled as an MLP with four hidden layers consisting of 256 hidden units. \cref{fig:model_architecture} shows the overview of our model architecture.
\vspace{-0.4em}
\subsection{Conditioning for Upsampling}
\vspace{-0.4em}
In the upsampling task, the model should be able to generate the details of the HR image while maintaining the low-frequency information of the given LR image well.
As conditioning information, we use the feature vector of LR image extracted by the auto-encoder $E_\varphi$.
At this time, the LR image is linearly interpolated to match the size of the target latent vector.
The conditioning information is strategically concatenated with intermediate feature maps at each denoising step. 
The LDM is learned to restore high-resolution image features from LR image features.

\vspace{-0.4em}
\subsection{Two-Stage Alignment Process}
\vspace{-0.4em}
The Latent Diffusion Model (LDM) is a two-stage model that has been designed to enhance the speed of both learning and inference, compared to diffusion models that operate in pixel space.
The second stage model relies on the intermediate expression or features extracted by the first stage model, an auto-encoder.
During the training process, the encoder can cause errors or inaccuracies, which are then passed on to the LDM.
Since the LDM also produces additional errors and is trained separately from the decoder, we anticipate that these errors would reduce the effectiveness of the decoding.
To address this issue,  we presented a two-stage alignment process to improve the quality of the output images by reducing misalignment errors between the two-stage models.

Ultimately, what we want is to generate the output image as similar as the ground truth image as possible.
Inspired by ~\cite{DiffusionCLIP,PASD}, we want to induce an image similar to ground-truth image to be generated through the loss between the result and the ground-truth image.
However, the larger the time step \(t \), the more difficult it is to accurately predict \(\hat{\text{z}}_0 \), so weights were given according to the time step \(t \). The reconstruction loss with the ground truth image is defined as
\vspace{-0.5em}
\begin{equation}
\vspace{-0.5em}
\begin{split}
  L_{recon} = \frac{\bar{\alpha}_t}{1-\bar{\alpha}_t} \left\| \text{x}_0 - \hat{\text{x}}_0 \right\|_{2}^{2}
  \label{eq:recon_loss}
\end{split}
\end{equation}
where \( \hat{\text{x}}_0 \) is generated image from the predicted latent \( \hat{\text{z}}_0 \). By combining \cref{eq:ldm_forward} and \cref{eq:ldm_loss} in \cref{sec:preliminary}, the denosing objective of LDM can be expressed as
\vspace{-0.5em}
\begin{equation}
\vspace{-0.5em}
\small
\begin{split}
  L_{dm} &\ = \left\| \epsilon - \epsilon_\theta\left( \text{z}_t \right) \right\|_{2}^{2} \\
  &\ = \left\| \frac{1}{\sqrt{1-\bar{\alpha}_t}} \left( \text{z}_t - \sqrt{\bar{\alpha}_t}\text{z}_0 \right) - \frac{1}{\sqrt{1-\bar{\alpha}_t}} \left( \text{z}_t - \sqrt{\bar{\alpha}_t}\hat{\text{z}}_0 \right) \right\|_{2}^{2} \\
  &\ = \frac{\bar{\alpha}_t}{1-\bar{\alpha}_t} \left\| \text{z}_0 - \hat{\text{z}}_0 \right\|_{2}^{2}
  \label{eq:ldm_loss_modified}
\end{split}
\end{equation}
Specifically, to fine-tune the diffusion model, we modified the objective function as:
\vspace{-0.5em}
\begin{equation}
\vspace{-0.5em}
\begin{split}
  L_{align} &\ = \lambda_{1} L_{dm}  + \lambda_{2} L_{recon}
  \label{eq:alignment_loss}
\end{split}
\end{equation}
The denoising network is fine-tuned at each reverse step in a similar way to the training strategy proposed by Kim \etal~\cite{DiffusionCLIP}.
We also adopted reverse DDIM~\cite{DDIM} process for fast sampling.
The overall flow of the proposed alignment process is shown in \cref{fig:model_architecture}.

\section{Experiment}
\label{sec:experiment}

\subsection{Implementation}
To train the implicit neural decoder and diffusion model, we use Adam optimizer with learning rates of 5e-5 and 1e-6, respectively.
We set both $\lambda_1$ and $\lambda_2$ to 1.0.
We utilized a single 24GB NVIDIA RTX 4090 GPU for all experiments.

\subsection{Evaluation}
\textbf{Datasets.}
We evaluate using the datasets below:

\begin{itemize}
    \item The Human Face Dataset contains two sub-datasets: Flick-Faces-HQ (FFHQ)~\cite{FFHQ} and CelebA-HQ~\cite{ProGAN}. These datasets consist of 70K and 30K different human face images, respectively.
    \item We used LSUN~\cite{LSUN} for general scenes. The LSUN database is divided into various subcategory images, with the smaller size of image is 256x256 pixels.
    \item To demonstrate the upsampling potential of our model on ultra-high-resolution images, we used the wild datasets DIV2K and Flickr2K.
\end{itemize}
The image resolutions used for comparison methods are different. For more detailed setup of the training dataset, see \cref{sec:Experiment_Datails} of the supplementary material.

\textbf{Metrics.}
We use evaluation metrics commonly used in Image Generation and Super-Resolution tasks.
For image generation task, Fréchet Inception Distance(FID), Precision, and Recall are used to measure the perceptual quality and reliability of the generated image and to measure how well it covers the distribution range of real images, respectively. 
In addition, SelfSSIM~\cite{ScaleParty} is also calculated to measure the consistency between images generated at different scales. 

In SR task, PSNR is used to compare how close the upscale is to the ground truth.
However, since PSNR does not capture high texture details well, it is known to not correlate well with human perception of image quality~\cite{SRGan, Trade-off, LPIPS}.
Therefore, we also use LPIPS~\cite{LPIPS} to compare perceptual quality with higher precision.
Finally, FPS (Frames Per Second) is measured for inference speed comparison.

\subsection{Comparisons on Image-Generation}
\begin{figure}[t]
  \centering
   \includegraphics[width=1.0\linewidth]{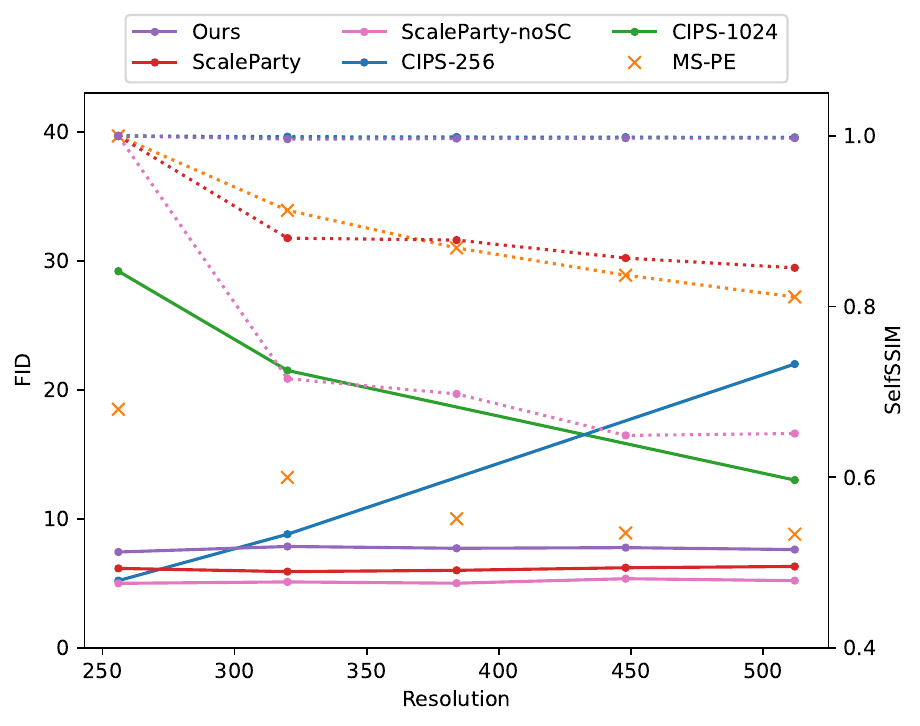}
   \vspace{-2.2em}
   \caption{We compare our model quantitatively with FID and SelfSSIM scores on the FFHQ datasets.
   The solid lines represent the FID scores of the methods that generate images of arbitrary-scale, while the dotted lines indicate the SelfSSIM scores.
   The `$\times$' symbol indicates the method that only generates images of a fixed scale.
   Our model demonstrates competitive performance in both evaluation metrics.
   }
   \vspace{-0.3em}
   \label{fig:ffhq_gen_comparison}
\end{figure}

\textbf{Quantitative Comparisons.}
For the qualitative comparison of arbitrary-scale image generation performance with other methods, we use arbitrary-scale generative models, CIPS~\cite{CIPS} and ScaleParty~\cite{ScaleParty}, and predefined multi-scale generative modes, MSPIE~\cite{MSPIE} and MS-PE~\cite{MS-PE}.
\cref{fig:ffhq_gen_comparison} and \cref{tab:quantitative_gen_lsun} show the qualitative comparison on FFHQ and LSUN Bedroom datasets, respectively.
For evaluation, 50k images were sampled.

CIPS, an INR-based model, has very high consistency between images generated at each scale.
However, the further away from the resolution is used for training, the lower the FID score is obtained.
ScaleParty guarantees good FID scores regardless of scales.
Although it also achieved a high score in terms of consistency, it clearly falls short compared to the INR-based model.
Our model not only achieves good FID scores on all scales, but also shows high consistency.
Moreover, our model shows much better diversity (recall) than other models, see \cref{tab:quantitative_gen_lsun}.

\begin{table}
  \centering
  \footnotesize
  \begin{tabular}{c|c|ccc|ccc}
    \toprule
    Dataset:    & \multicolumn{7}{c}{LSUN Bedroom}                                \\
    \midrule
    Method      & Res & FID$\downarrow$   & Prec$\uparrow$  & Rec$\uparrow$   & \multicolumn{3}{c}{SelfSSIM (5k)$\uparrow$} \\
    \midrule
    MSPIE       & 128& 11.39 & \textbf{66.45} & 26.97 & 1.00 & 0.10 & 0.10      \\
                & 160 & 16.45 & 63.84 & 23.09 & 0.10 & 1.00 & 0.12      \\
                & 192 & 12.65 & 58.10 & 25.93 & 0.10 & 0.12 & 1.00      \\
    Scaleparty  & 128 & 10.15 & 62.50 & 20.63 & 1.00 & 0.94 & 0.92      \\
                & 160 & 9.85  & \textbf{64.14} & 22.02 & 0.92 & 1.00 & 0.95      \\
                & 192 & 9.91  & \textbf{64.77} & 21.10 & 0.89 & 0.94 & 1.00      \\
    Ours        & 128 & \textbf{7.20}  & 59.69 & \textbf{38.26} & 1.00 & \textbf{0.98} & \textbf{0.99}      \\
                & 160 & \textbf{7.43}  & 58.52 & \textbf{32.12} & \textbf{0.96} & 1.00 & \textbf{0.99}      \\
                & 192 & \textbf{7.73}  & 59.57 & \textbf{27.98} & \textbf{0.95} & \textbf{0.97} & 1.00     \\
    \bottomrule
  \end{tabular}
  \vspace{-1.0em}
  \caption{Comparison of quantitative results on LSUN Bedroom datasets. For more results on other dataset, see \cref{tab:quantitative_gen_church} in the supplementary material.}
  \vspace{-1.3em}
  \label{tab:quantitative_gen_lsun}
\end{table}

\begin{figure}[t!]
  \centering
   \includegraphics[width=1.0\linewidth]{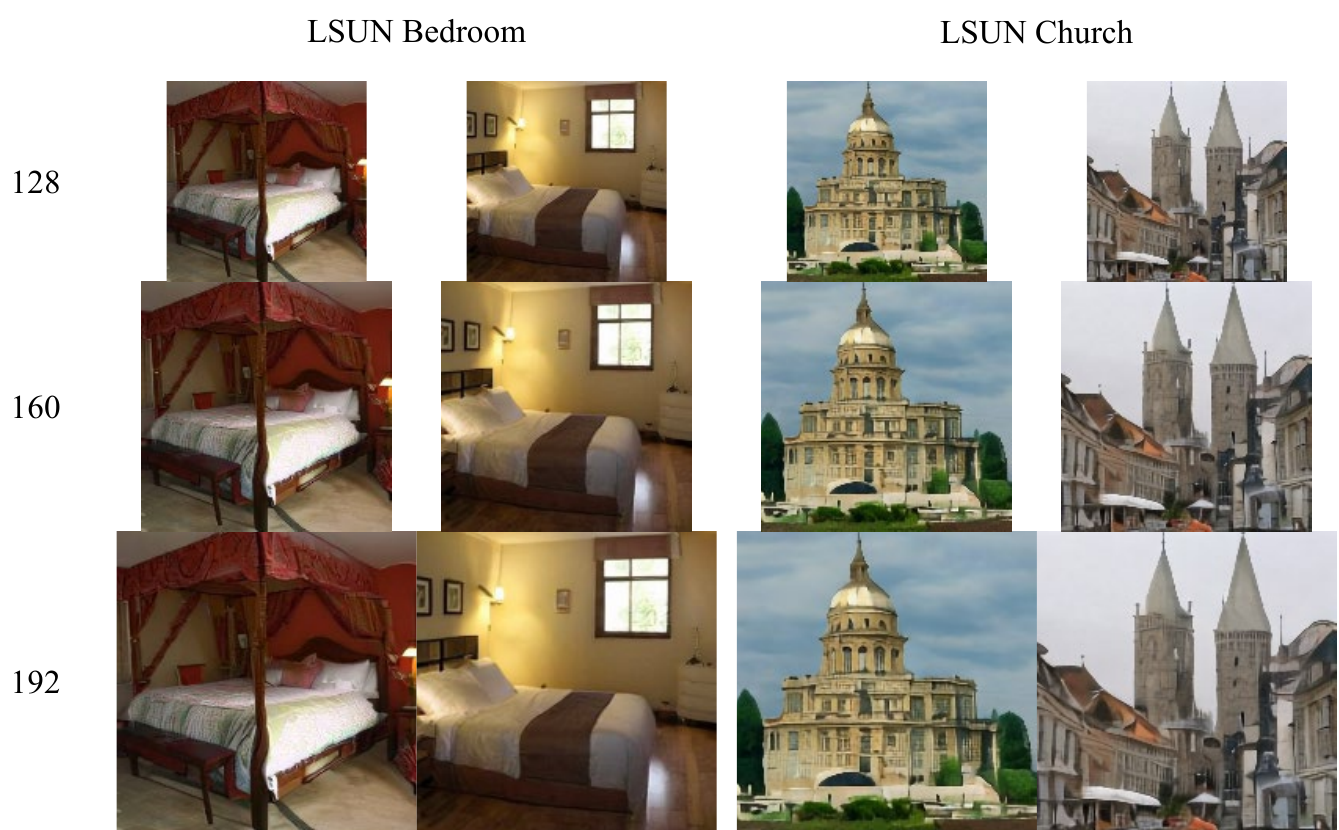}
  \vspace{-1.8em}
   \caption{Qualitative results on LSUN Bedroom and Church datasets.
   For more results, see supplementary.}
  \vspace{-0.8em}
   \label{fig:lsun_gen}
\end{figure}

\textbf{Visualization.}
The qualitative results of our model is illustrated in \cref{fig:teaser} and \cref{fig:lsun_gen}. We selected some arbitrary scales and visualized the results.
We can observe the model's effectiveness in generating diverse images, with high perceptual quality on various datasets. 
It also shows high scale-consistency on different scales.
See Supplementary for more qualitative results.
\begin{figure}[t!]
  \centering
   \includegraphics[width=1.0\linewidth]{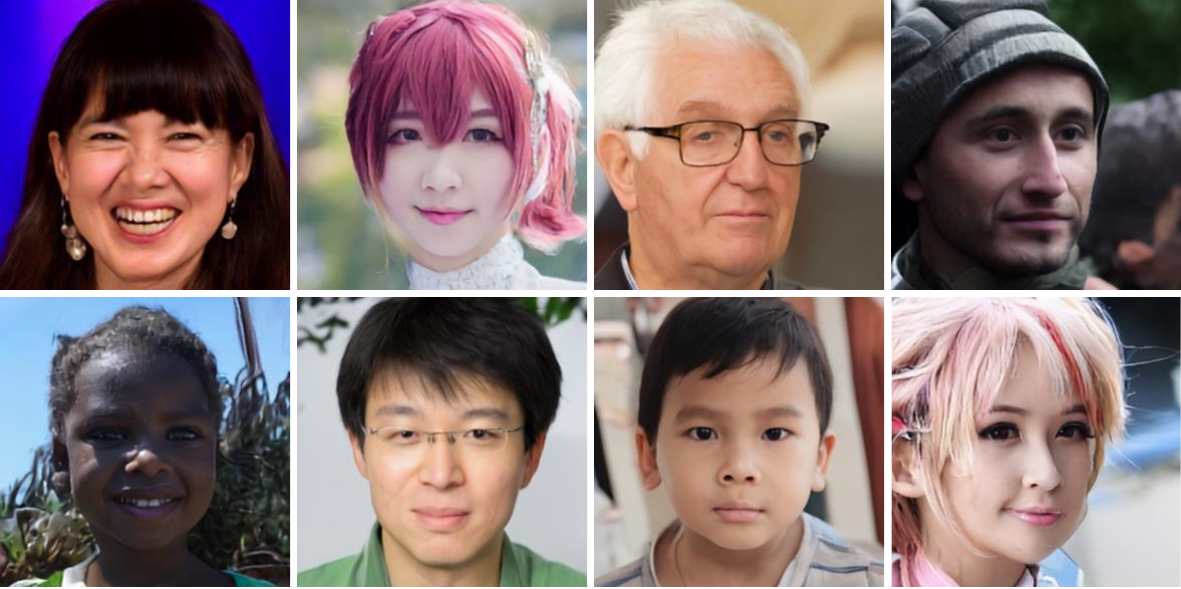}
  \vspace{-1.8em}
   \caption{Various generated images on face dataset.
   See \cref{fig:ffhq_gen_dirversity} in the supplementary to see larger and more diverse images.}
  \vspace{-1.3em}
   \label{fig:ffhq_gen_diversity}
\end{figure}
\begin{table*}[ht]
  \fontsize{7.8pt}{7.8pt}\selectfont
  \centering
  \begin{tabular}{c|cc|ccc}
    \toprule
    Dataset:        & \multicolumn{5}{c}{CelebA-HQ} \\
    \midrule
    Method             & 5.3$\times$     & 7$\times$         & 10$\times$    & 10.7$\times$  & 12$\times$     \\
    \midrule
    LIIF~\cite{LIIF}   & \textbf{27.52} / 0.1207   & \textbf{25.09} / 0.1678   & 22.97 / 0.2246 & 22.39 / 0.2276 & 21.81 / 0.2332 \\
    SR3~\cite{SR3}     & -                & 21.15 / 0.1680   & 20.25 / 0.2856 & -              & 19.48 / 0.3947 \\
    IDM~\cite{IDM}     & 23.34 / 0.0526   & 23.55 / 0.0736   & 23.46 / 0.1171 & 23.30 / 0.1238 & 23.06 / 0.1800 \\
    Ours               & 24.66 / \textbf{0.0455}   & 24.13 / \textbf{0.0690}   & \textbf{23.64 / 0.1110} & \textbf{23.62 / 0.1183} & \textbf{23.52 / 0.1427} \\
    \bottomrule
  \end{tabular}
  \hfill
  \begin{tabular}{c|c|c}
    \toprule
    Dataset:           & Lsun Bedroom               & LSUN Tower \\
    \midrule
    Method             & \multicolumn{2}{c}{16$\times$}          \\
    \midrule
    PULSE~\cite{PULSE} & 12.97 / 0.7131             & 13.62 / 0.7066 \\
    GLEAN~\cite{GLEAN} & 19.44 / 0.3310             & 18.41 / 0.2850 \\
    IDM~\cite{IDM}     & \textbf{20.33} / 0.3290    & 19.44 / 0.2549 \\
    Ours               & 20.08 / \textbf{0.3269}    & \textbf{21.24} / \textbf{0.1897} \\
    \bottomrule
  \end{tabular}
  \vspace{-1.0em}
  \caption{Quantitative results of arbitrary-scale super-resolution on CelebA-HQ and LSUN Bedroom datasets. For each method, PSNR$\uparrow$/LPIPS$\downarrow$ scores are reported.}
  \vspace{-1.6em}
  \label{tab:face_sr}
\end{table*}

\subsection{Comparisons on Arbitrary SR}
\vspace{-0.5em}
\textbf{Quantitative Comparisons.}
Following IDM, we evaluated our model on CelebA-HQ face images.
\cref{tab:face_sr} shows the quantitative results with LIIF, SR3, and IDM.
LIIF and IDM, as well as our model, are arbitrary scale super-resolution models, trained in the scale range (1, 8] using FFHQ datasets.
Although SR3 is a model that operates only at a fixed integer magnification, it shows a lower PSNR and higher LPIPS compared to other methods.
LIIF shows good PSNR scores for in-distribution scales, but still shows low perceptual quality.
IDM achieved a low LPIPS score while maintaining a high PSNR.
Except for the in-distribution PNSR score of LIIF, our model outperformed other methods in PSNR and LPIPS at all scales.

Similar results were obtained with the in-the-wild dataset (see \cref{tab:re_div_sr} and \cref{tab:re_div_out}).
Our model shows superior performance compared to other generative models despite using less training data.
Although regression-based methods like EDSR and LIIF have shown high scores on 4$\times$, \cref{tab:re_div_out} indicates that our approach surpasses LIIF at larger scale.
Additionally, our model, based on diffusion models, offers the benefits of diversity and scale consistency.

\begin{table}[t]
  \centering
  \small
    \begin{tabular}{cl|c|cc}
    \toprule
    \multicolumn{2}{c|}{Method}  & Datasets & PSNR$\uparrow$  & SSIM$\uparrow$ \\
    \midrule
    Reg.-based    & EDSR        & D+F      & 28.98 & 0.83 \\
                  & LIIF        & D+F      & \textbf{29.00} & \textbf{0.89} \\
    \midrule
    \midrule
    GAN-based     & ESRGAN      & D+F      & 26.22 & 0.75 \\
                  & RankSRGAN   & D+F      & 26.55 & 0.75 \\
    \midrule
    Flow-based    & SRFlow      & D+F      & 27.09 & 0.76 \\
    \midrule
    Flow+GAN      & HCFlow++    & D+F      & 26.61 & 0.74 \\
    \midrule
    Diffusion     & IDM         & D        & 27.10 & 0.77 \\
                  & IDM         & D+F      & 27.59 & 0.78 \\
                  & Ours        & D        & \textbf{27.61}      & \textbf{0.81}   \\
    \bottomrule
    \end{tabular}
    \vspace{-1.0em}
    \caption{Quantitative comparison of 4$\times$ super-resolution using in-the-wild datasets. D and F refer to the DIV2k and Flickr2k.
    }
    \vspace{-1.2em}
  \label{tab:re_div_sr}
\end{table}

\begin{table}
  \centering
  \small
  \begin{tabular}{c|ccc}
    \toprule
    Method                      & 8$\times$                 & 12$\times$                & 17$\times$        \\
    \midrule
    LIIF & \textbf{23.97} / 0.4790   & 22.28 / 0.5900            & 21.23 / 0.6560     \\ 
    Ours                        & 23.82 \textbf{/ 0.4265}   & \textbf{22.73 / 0.5463}   & \textbf{21.83 / 0.6225}        \\
    \bottomrule
  \end{tabular}
  \vspace{-1.0em}
  \caption{Comparison (PSNR$\uparrow$ / LPIPS$\downarrow$) on the DIV2K at out-of-distribution scales.}
  \vspace{-1.0em}
  \label{tab:re_div_out}
\end{table}

\begin{figure}[ht]
  \centering
  \begin{subfigure}{1.0\linewidth}
     \includegraphics[width=0.98\linewidth]{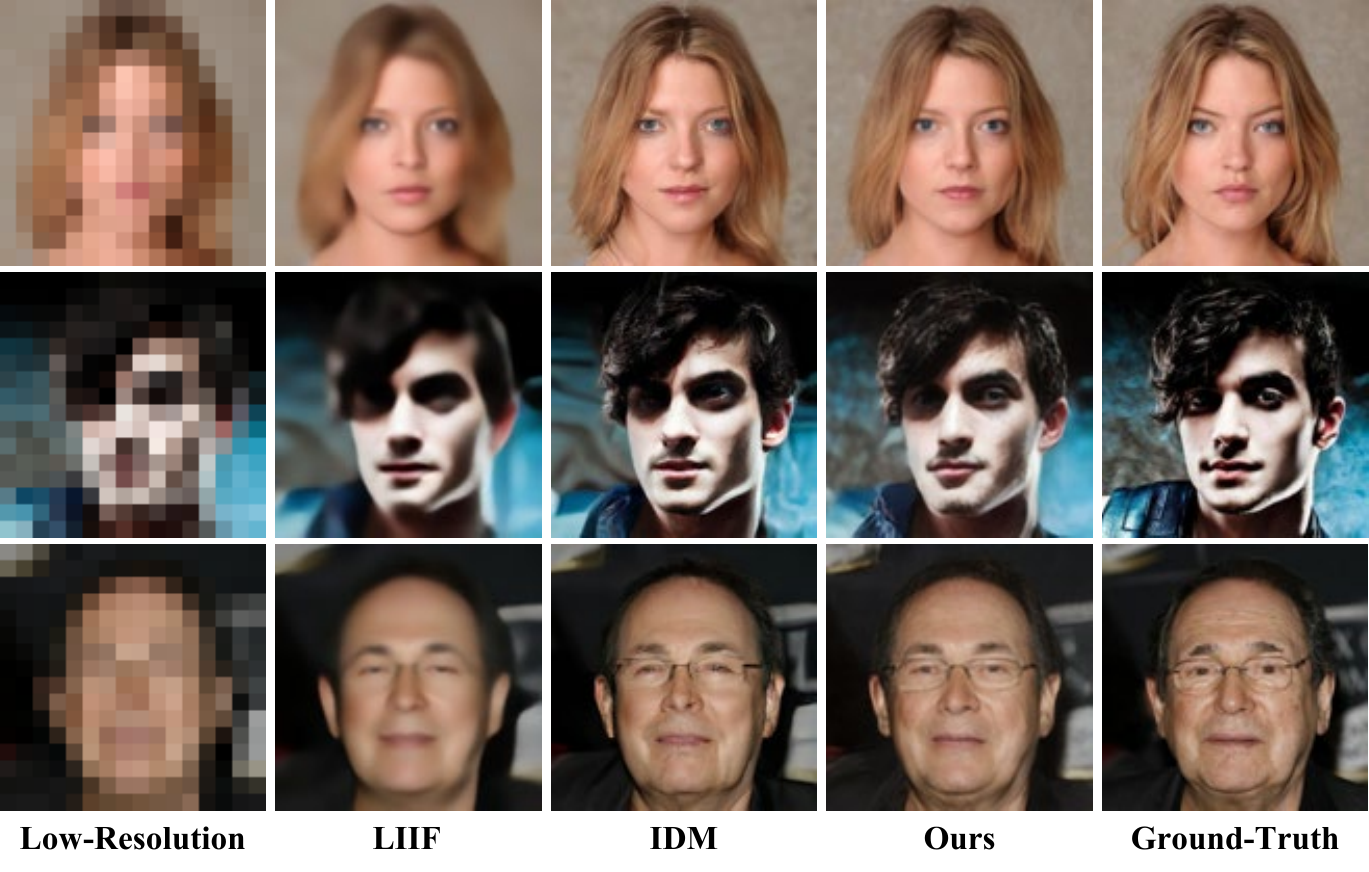}
  \end{subfigure}
  \vfill
  \begin{subfigure}{1.0\linewidth}
     \includegraphics[width=0.98\linewidth]{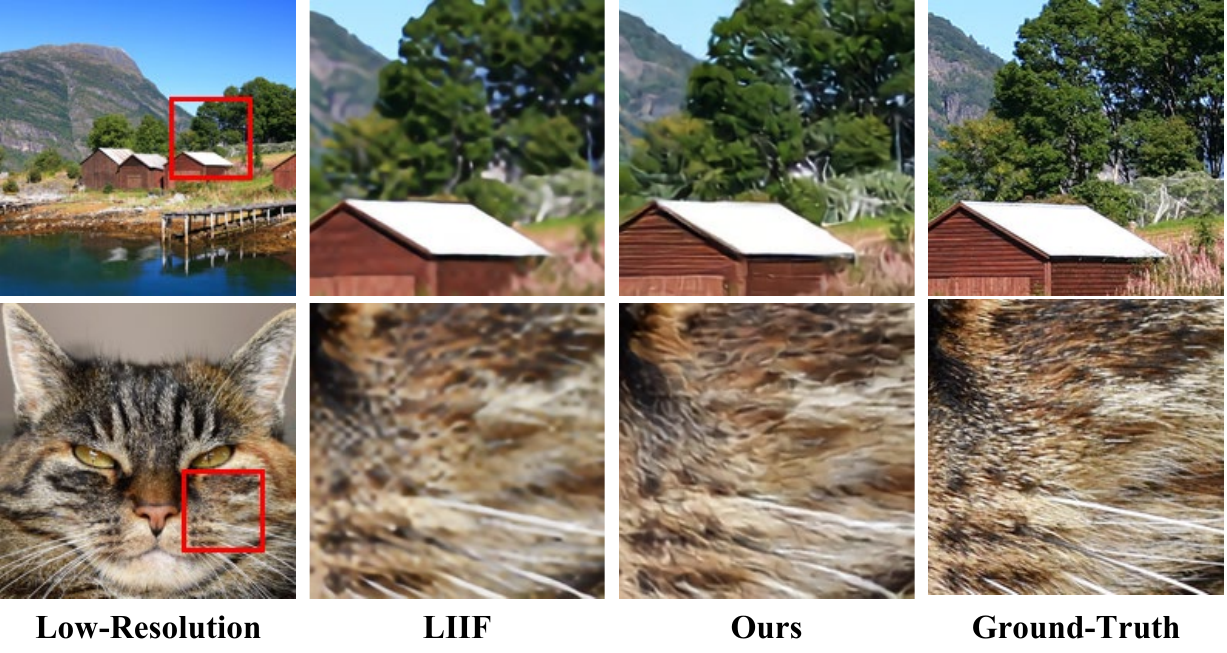}
  \end{subfigure}
  \vspace{-2.0em}
   \caption{Qualitative comparisons of arbitrary-scale upsampling on face (\textbf{upper}) and in-the-wild (\textbf{lower}) dataset.
   Our model restored more detailed information and generated faces with similar expressions to ground-truth images compared to other models.}
  \vspace{-1.4em}
   \label{fig:qualitative_sr}
\end{figure}

\begin{figure*}[t!]
  \centering
   \includegraphics[width=0.59\linewidth]{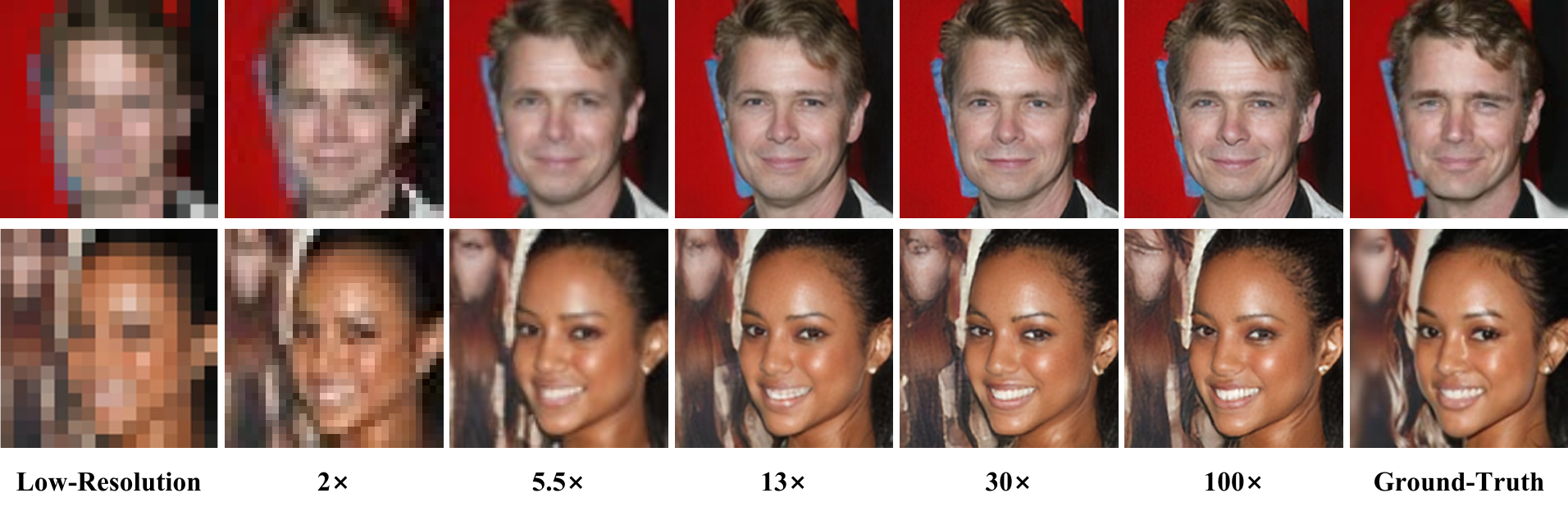}
   \hfill
   \includegraphics[width=0.39\linewidth]{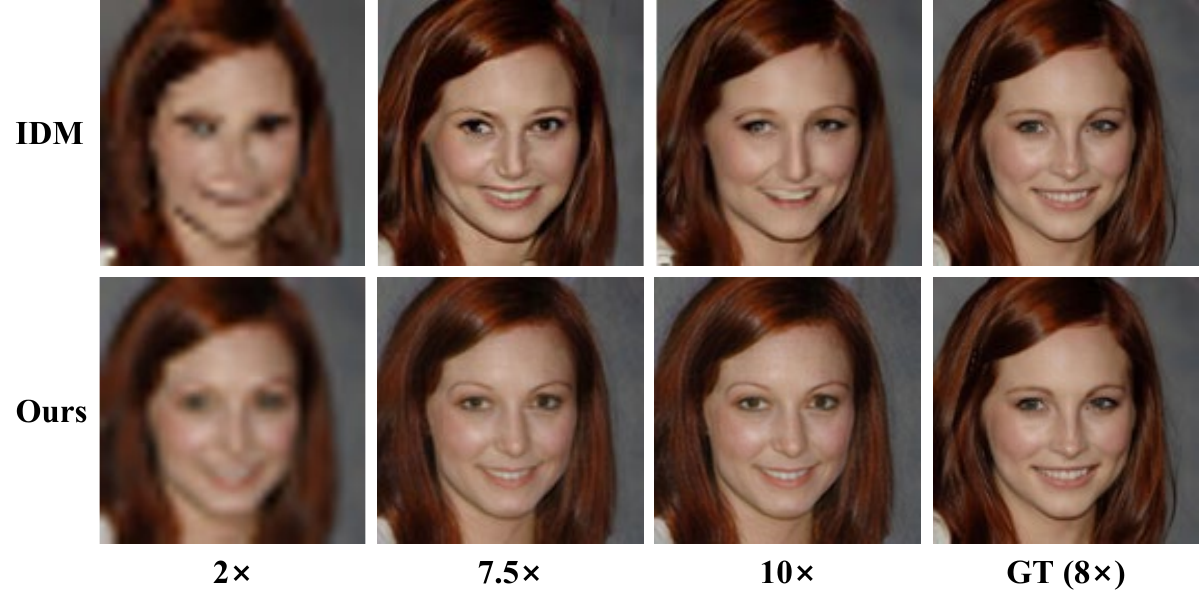}
  \vspace{-1.0em}
   \caption{\textbf{Left:} Qualitative results of the proposed method for arbitrary-scale upsampling on face datasets.
   \textbf{Right:} Comparison of scale consistency on face dataset.}
  \vspace{-1.5em}
   \label{fig:qualitative_sr_arbitrary}
\end{figure*}

\textbf{Visualization.}
We visualize the upsampling results of several face datasets to demonstrate  qualitative comparisons on arbitrary-scale upsampling in \cref{fig:qualitative_sr}.
A very low-resolution image does not contain enough features, making it very difficult to up-sample.
Despite these difficulties, LIIF restores a face with an expression very similar to that of the ground-truth image.
However, the output image is over-smoothed, and detailed textures such as hair are not sufficiently restored.
IDM generates images very realistically and delicately.
Although high consistency is maintained across scales, differences in facial expressions are still visible.
In contrast, our model not only maintains high fidelity and perceptual quality, but also produces consistent images well without significant distortion even at arbitrary large scales, see \cref{fig:qualitative_sr_arbitrary}.

\begin{figure}[t!]
  \centering
  \includegraphics[width=1.0\linewidth]{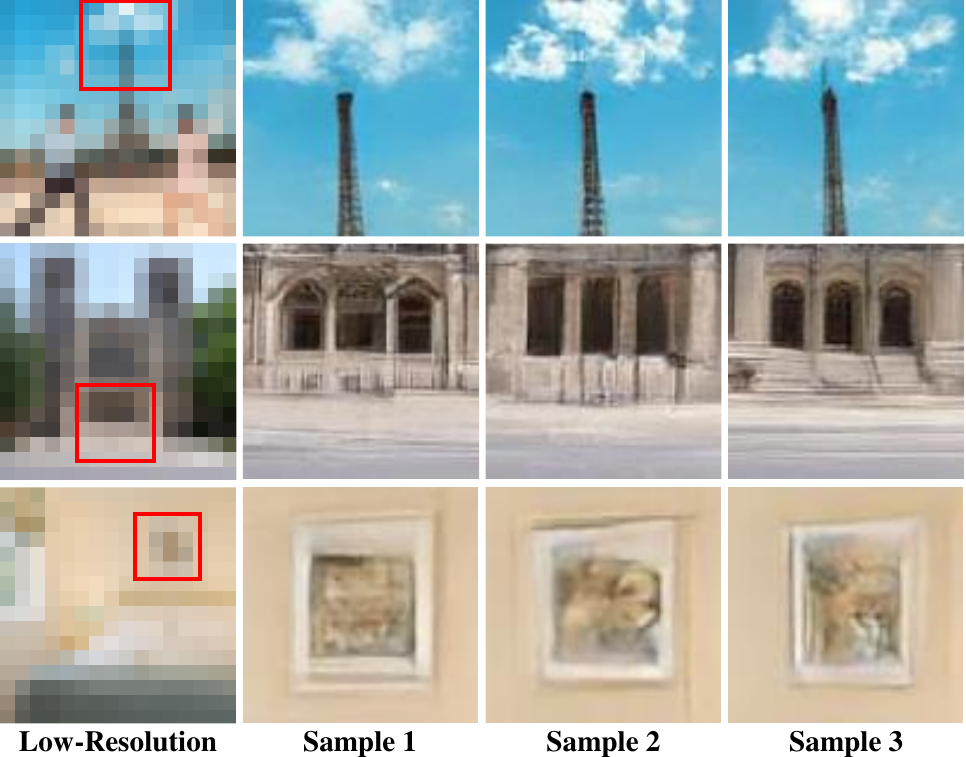}
  \vspace{-1.5em}
   \caption{Visualization of result diversity in super-resolution tasks.
   This allows our model to better handle the `ill-posed problem'.
   See \cref{fig:sr_diversity} in the supplementary to see larger and more diverse results.}
  \vspace{-0.5em}
   \label{fig:lsun_sr_diversity}
\end{figure}

\textbf{Inference Speed and Memory Usage.}
Given that diffusion models typically suffer from slow inference speeds and high memory usage compared to other methods, we have intentionally designed our model to alleviate these shortcomings.
We compared the inference speed of each model by measuring FPS (Frames Per Second) and \cref{tab:quantitative_speed_memory} shows the results.
MLP-based implicit networks require a long time compared to other network structures.
In IDM, an implicit network must be passed at every denoising step, which leads to significant slowdown.
Additionally, since a high-dimensional denoising process is required to generate a large image, the larger the scale is, the more severely the inference speed decreases.
In contrast, our model compensates the speed by using the implicit network only once during the decoding process.
As a result, our model is able to inference faster than IDM (about 12.7 times faster on 8$\times$ task), while showing better output quality.
Furthermore, since IDM operates in the pixel space, it was not possible to generate very large-scale images such as 200$\times$ in our environment (Single RTX 4090), but our model can generate them by adjusting the input batch size of the MLP.
Our model is still slower than other GAN-based SR models or regression models, but it shows a fast inference speed compared to other diffusion-based models.

\begin{table}
  \centering
  \small
  \begin{tabular}{c|ccccc}
    \toprule
  Method &   \multicolumn{5}{c}{FPS$\uparrow$}                          \\
  & 8$\times$ & 12$\times$ & 30$\times$ & 100$\times$ & 200$\times$ \\
    \midrule
  IDM & 0.0202 & 0.0200 & 6.54e-3 & 3.98e-4 & - \\
  Ours & \textbf{0.2568} & \textbf{0.2510} & \textbf{0.2473} & \textbf{0.1833} & \textbf{0.0982} \\
    \bottomrule
  \end{tabular}
  \vspace{-0.8em}
  \caption{Comparison of inference Speed in terms of FPS (Frames Per Second). `-' indicates that the model did not work in our environment due to memory overflow.}
  \vspace{-0.8em}
  \label{tab:quantitative_speed_memory}
\end{table}

\begin{table}
  \centering
  \begin{tabular}{c|cccc}
    \toprule
                                        & \multicolumn{4}{c}{\textbf{Encoder-Decoder (PSNR$\uparrow$)}} \\
                                        & \multicolumn{2}{c}{16 \(\rightarrow\) 128 (\text{8}$\times$)}  & \multicolumn{2}{c}{64 \(\rightarrow\) 256 (\text{4}$\times$)}  \\
    \midrule
    MLP    & \multicolumn{2}{c}{22.96}   & \multicolumn{2}{c}{32.54 }     \\
    AE+MLP & \multicolumn{2}{c}{\textbf{35.28}}   & \multicolumn{2}{c}{\textbf{35.93}}      \\
    \midrule
    \midrule
                       & \multicolumn{4}{c}{\textbf{Aligment Process (PSNR$\uparrow$ / LPIPS$\downarrow$)}} \\
                       & 5.3$\times$     & 7$\times$       & 10$\times$     & 12$\times$     \\
    \midrule
    \multirow{2}{*}{Before} &   22.79 /      &      22.49 /     &   23.23 / & 22.14 / \\
                            &   0.0600      &      0.0932     &   0.1691 & 0.2077 \\
    \multirow{2}{*}{After}  &   \textbf{24.66 /}   & \textbf{24.13 /}   & \textbf{23.64 /}  & \textbf{23.52 /}  \\
                            &   \textbf{0.0455}   & \textbf{0.0690}   & \textbf{0.1110}  & \textbf{0.1427}  \\
    \bottomrule
  \end{tabular}
  \vspace{-0.8em}
  \caption{Ablation studies of decoder structures and alignment process.}
  \vspace{-0.8em}
  \label{tab:ablation}
\end{table}

\subsection{Ablation Studies}
\vspace{-0.5em}
\textbf{Decoder Architecture.}
\cref{tab:ablation} shows the reconstruction capabilities according to the decoder structures.
Rather than directly reconstructing the extracted features using MLPs, its reconstruction quality is better when using the symmetric decoder from the pre-trained auto-encoder.

\begin{figure}
  \centering
   \includegraphics[width=0.95\linewidth]{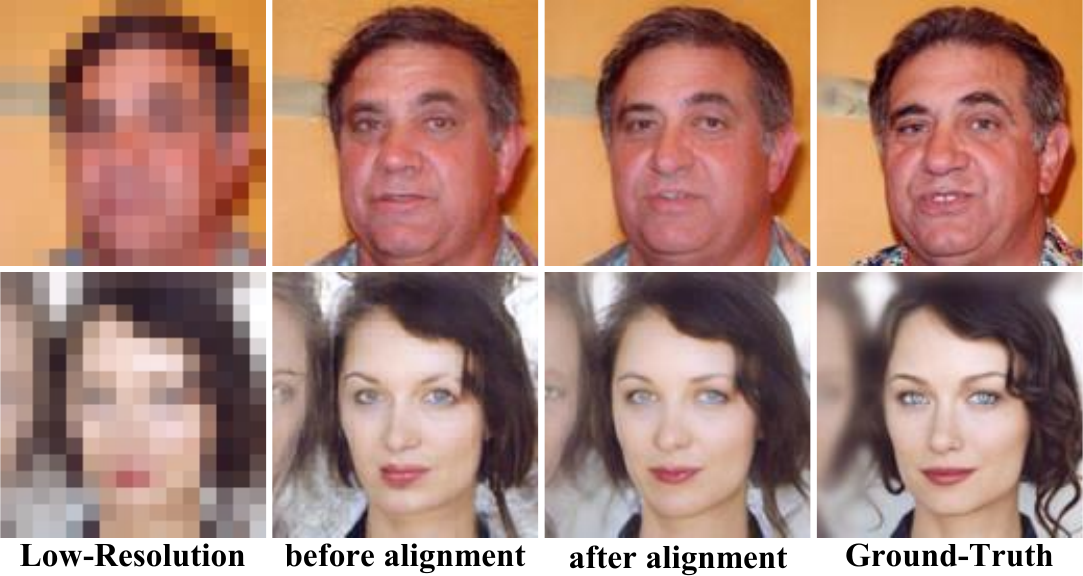}
  \vspace{-1.0em}
   \caption{Comparison of qualitative results before and after the two-stage alignment process.}
   \vspace{-1.3em}
   \label{fig:ablation}
\end{figure}

\textbf{Alignment Process.}
\cref{tab:ablation} and \cref{fig:ablation} present the quantitative and qualitative results of the two-stage alignment process, respectively.
To demonstrate the effectiveness of the two-stage alignment process, we compare the results before and after the process.
Overall, artifacts are reduced and textures are more realistic.

\vspace{-0.8em}
\section{Conclusions}
\vspace{-0.6em}
\label{sec:conclusion}
In this paper, we proposed an Implicit Neural Decoder with a latent diffusion model for arbitrary-scale image generation and upsampling.
We show that our Implicit Neural Decoder can effectively reconstruct latent features into continuous-scale images.
This has enabled our model to train and infer efficiently in the latent space.
As a result, our method is an order-of-magnitude faster compared to IDM.
Furthermore, we also proposed a two-stage alignment process to improve the quality of output images by reducing misalignment errors between the two-stage models.
We achieved high-diversity image generation and high-fidelity super-resolution at arbitrary scales while maintaining perceptual quality and scale consistency. 

\textbf{Acknowledgements.}
This work was in part supported by NST grant (CRC 21011, MSIT), KOCCA grant (R2022020028, MCST), IITP grant (RS-2023-00228996, MSIT), and LG Electronics Co., Ltd.

{
    \small
    \bibliographystyle{ieeenat_fullname}
    \bibliography{main}
}


\clearpage
\setcounter{page}{1}
\setcounter{section}{0}
\setcounter{figure}{0}
\setcounter{table}{0}
\renewcommand{\thesection}{S.\arabic{section}}
\renewcommand{\thetable}{S\arabic{table}}
\renewcommand{\thefigure}{S\arabic{figure}}
\maketitlesupplementary

In this Supplementary Material, we describe the evaluation metrics in more details. In addition, we present more qualitative results for comparing other models and visualizing the quality of output images.

\section{Metrics}
\label{sec:Metrics}
\textbf{FID~\cite{suppl_FID}.}
Fréchet inception distance is a metric used for evaluating the quality of generated images produced by generative models.
It measures the similarity between the distribution of real images and the distribution of generated images by computing the Fréchet distance in feature space.

\textbf{Precision and Recall~\cite{suppl_PreRec}.}
Precision and recall are proposed metrics to evaluate fidelity and diversity.
Precision refers to the ratio of the generated image to the real image distribution and refers to the precision of how accurately the generated image depicts the real image sample. Recall refers to the ratio of the actual image to the distribution of the generated image sample and refers to the diversity of the generated image.

\textbf{SelfSSIM~\cite{suppl_ScaleParty}.}
It is a metric for evaluating the scale consistency of the generated images.
We downsample two images of different resolutions to a lower resolution and then measure the SSIM~\cite{suppl_SSIM} between them.

\textbf{PSNR.}
Peak Signal-to-Noise Ratio (PSNR) is a widely used metric to quantify the quality of a reconstructed image compared to the original image.
A higher PSNR score indicates a lower distortion.
It suggests that the processed image is closer to the original in terms of pixel-wise similarity.
However, recent research shows that this metric has limitations in indicating actual perceptual quality~\cite{suppl_Trade-off, suppl_SRGan, suppl_LPIPS}.

\textbf{LPIPS~\cite{suppl_LPIPS}.}
Learned Perceptual Image Patch Similarity (LPIPS) is a metric created to measure the similarity between image patches from a perceptual standpoint.
It has been demonstrated to accurately reflect human perception.
A low LPIPS score indicates that the patches are perceptually similar.

\section{Experiment Details}
\label{sec:Experiment_Datails}
In this section, we describe the target scale at which our model and the comparison models were trained on each task of the experiment.

\subsection{Image-Generation}
\label{sec:Experiment_Datails_Gen}
For the results in the comparative experiments, we referred to the results of Ntavelis \etal~\cite{suppl_ScaleParty}. \textcolor{blue}{Note ScaleParty, MSPIE, MS-PE were trained at larger resolutions, e.g. over 256$\times$256 and 128$\times$128 for FFHQ and LSUN respectively, while our method was trained at less than those resolutions.}

\textbf{FFHQ~\cite{suppl_FFHQ}.}
For the human face generation task, each model generated images at five different scales, 256, 320, 384, 448 and 512.
The training policy for each model is as follows.
\begin{itemize}
    \item MS-PE is trained for every scale in comparison (\ie 256, 320, 384, 448 and 512), since it is a multi-scale generation model.
    \item CIPS is trained on single scale, 256.
    \item ScaleParty is trained with two different resolutions, 256 and 384, for its scale consistency approach.
    \item Our model is trained to generate an image of arbitrary resolution between (32, 256] from a latent vector of 32.
\end{itemize}

\textbf{LSUN~\cite{suppl_LSUN}.}
For generic scene (bedroom and church) generation tasks, each model generated images at three different scales, 128 160 and 192.
The training policy for each model is as follows.
\begin{itemize}
    \item MSPIE and ScaleParty are trained for 128 and 192.
    \item Our model is trained to generate an image of arbitrary resolution between (64, 128] from a latent vector of 64.
\end{itemize}

\subsection{Super-Resolution}
In the super-resolution operation, 16$\times$16 low-resolution images are upsampled to arbitrary scales.
All models were trained within a scale range of 8$\times$ for human faces and 16$\times$ for generic scenes.

\section{More Results}
\label{sec:More_Results}
\subsection{Quantitative Results}
\cref{tab:quantitative_gen_church} show additional quantitative results of image generation for LSUN Church.
\textcolor{blue}{In the generation task, the maximum resolution of the images used by our model for training is lower than that of other models, as mentioned in \cref{sec:Experiment_Datails_Gen}.
Nevertheless, as shown in \cref{fig:ffhq_gen_comparison,tab:quantitative_gen_lsun} of the main text and \cref{tab:quantitative_gen_church}, our model shows competitive results.
In particular, our model shows great strengths in terms of diversity and scale consistency.}
And in the super-resolution task, our model achieves significantly better performance not only in terms of fidelity but also in terms of perceptual quality. All methods were trained at the same scales for the super-resolution task.

\begin{table}
  \centering
  \footnotesize
  \begin{tabular}{c|c|ccc|ccc}
    \toprule
    Dataset:    & \multicolumn{7}{c}{LSUN Church}                                \\
    \midrule
    Method      & Res & FID$\downarrow$   & Prec$\uparrow$  & Rec$\uparrow$   & \multicolumn{3}{c}{SelfSSIM (5k)$\uparrow$} \\
    \midrule
    MSPIE       & 128 &  \textbf{6.67} & \textbf{71.95} & 44.59 & 1.00 & 0.32 & 0.43      \\
                & 160 & 10.76 & 66.21 & 36.95 & 0.31 & 1.00 & 0.40      \\
                & 192 &  \textbf{6.02} & 66.70 & 46.13 & 0.39 & 0.38 & 1.00      \\
    Scaleparty  & 128 &  9.08 & 70.52 & 39.93 & 1.00 & 0.95 & 0.93      \\
                & 160 &  \textbf{7.96} & \textbf{70.87} & 32.07 & 0.94 & 1.00 & 0.95      \\
                & 192 &  7.52 & \textbf{68.14} & 33.33 & 0.90 & 0.94 & 1.00      \\
    Ours        & 128 & 8.25  & 65.27 & \textbf{47.02} & 1.00 & \textbf{0.98} & \textbf{0.98}      \\
                & 160 & 8.58  & 64.02 & \textbf{43.04} & \textbf{0.97} & 1.00 & \textbf{0.99}      \\
                & 192 & 8.81  & 62.36 & \textbf{42.80} & \textbf{0.96} & \textbf{0.97} & 1.00     \\
    \bottomrule
  \end{tabular}
  \caption{Quantitative comparison of image generation on LSUN Church datasets.}
  \label{tab:quantitative_gen_church}
\end{table}

\subsection{Qualitative Results}
\begin{figure*}[t]
  \centering
  \includegraphics[width=0.9\linewidth]{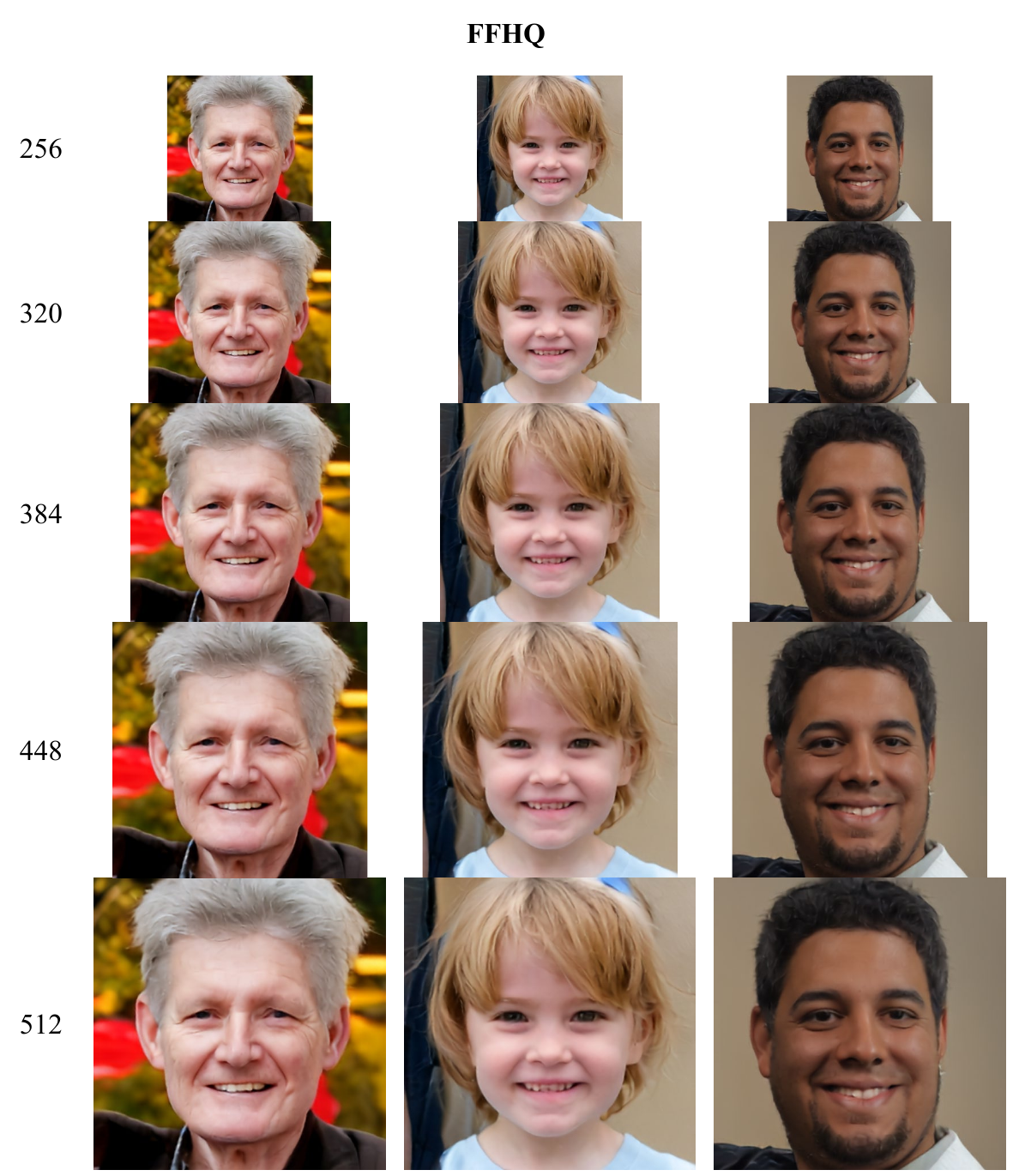}
  \caption{
  Scale consistency results of image generation on the FFHQ datasets.
  }
  \label{fig:ffhq_gen_scale}
\end{figure*}

\begin{figure*}[t]
  \centering
  \includegraphics[width=0.9\linewidth]{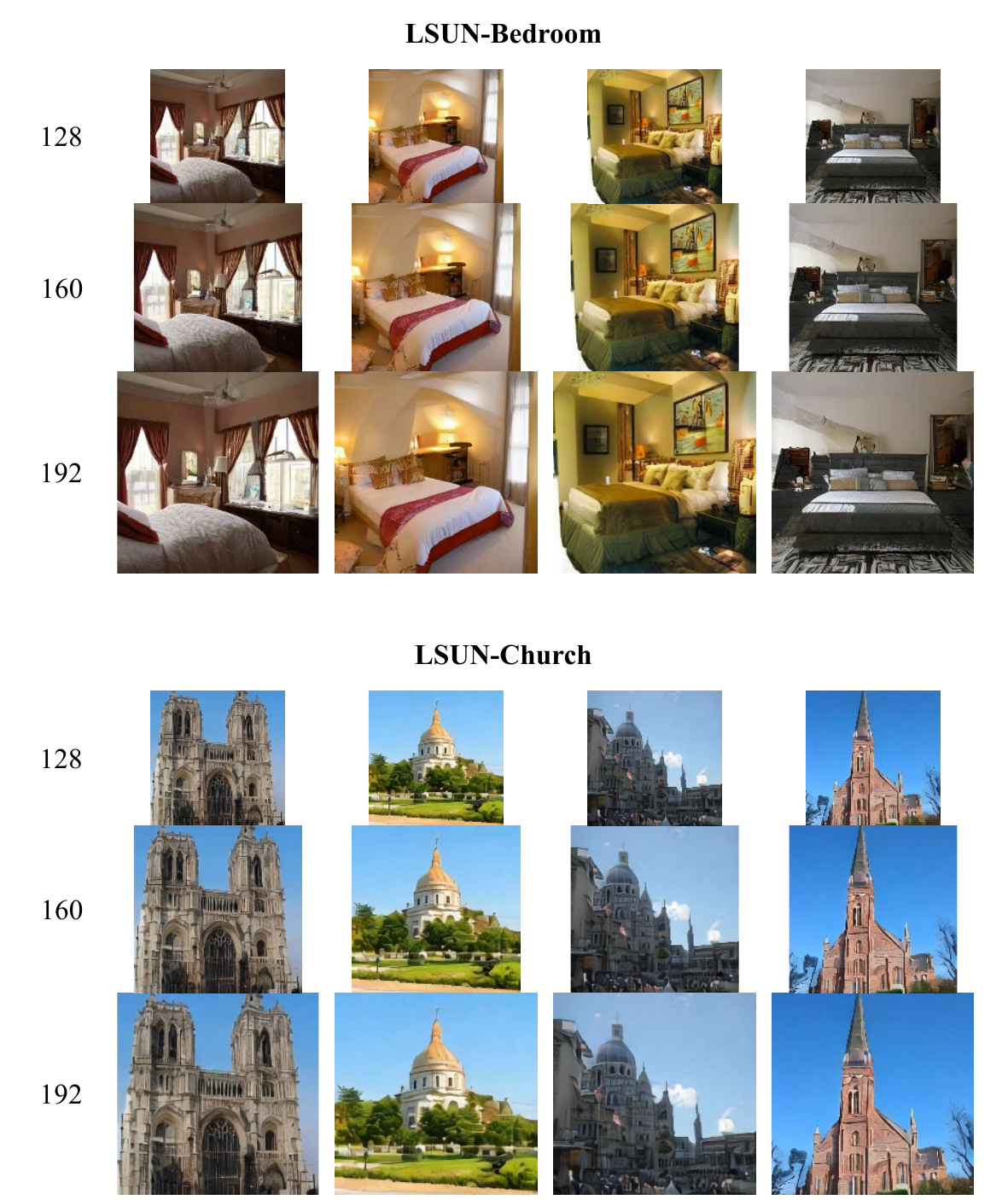}
  \caption{
  Scale consistency results of image generation on the LSUN Bedroom, Church datasets.
  }
  \label{fig:lsun_gen_scale}
\end{figure*}

\begin{figure*}[t]
  \centering
  \includegraphics[width=0.9\linewidth]{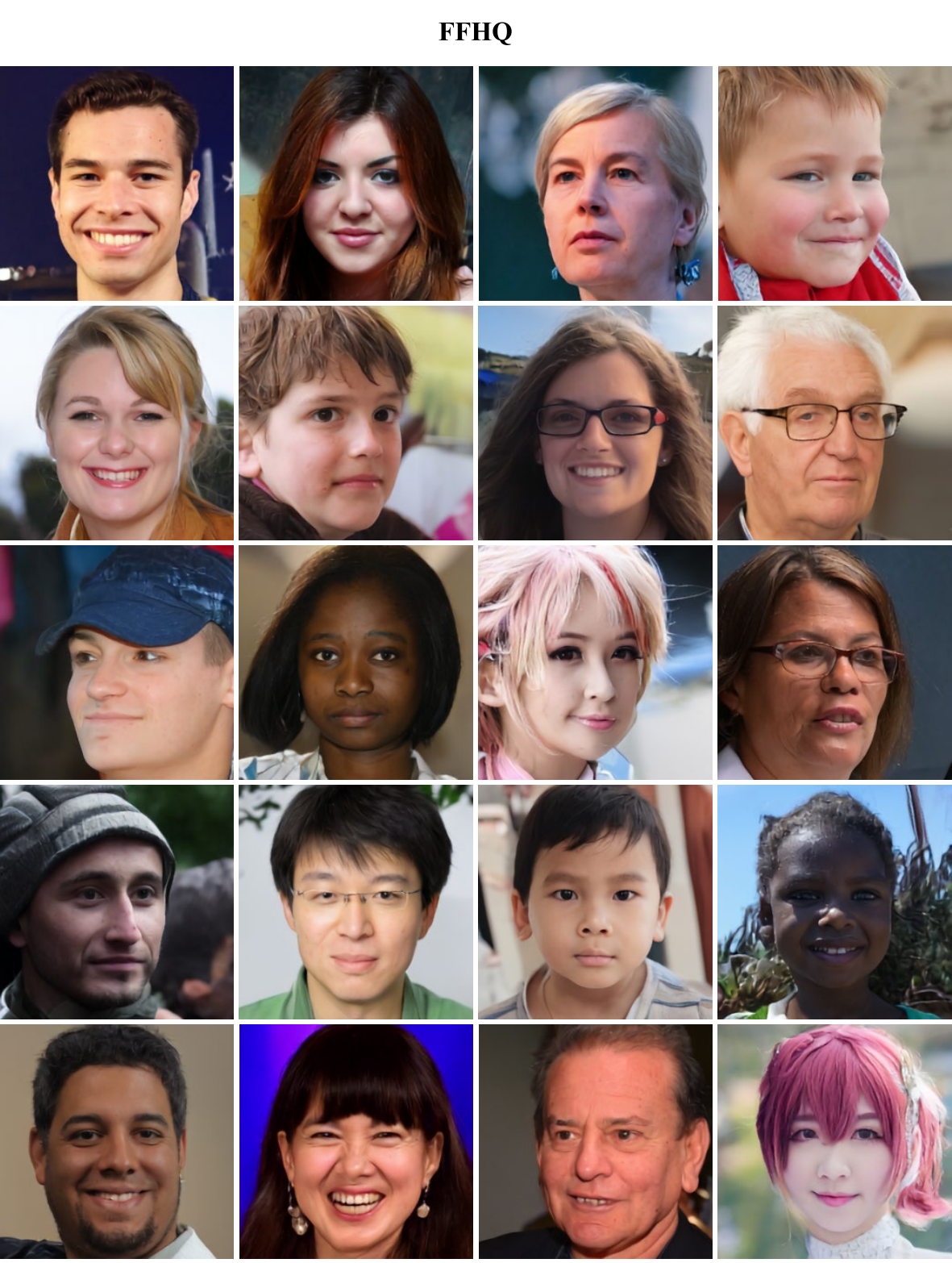}
  \caption{
 Visual results of image generation on the FFHQ datasets.
  }
  \label{fig:ffhq_gen_dirversity}
\end{figure*}

\begin{figure*}[t]
  \centering
  \includegraphics[width=0.9\linewidth]{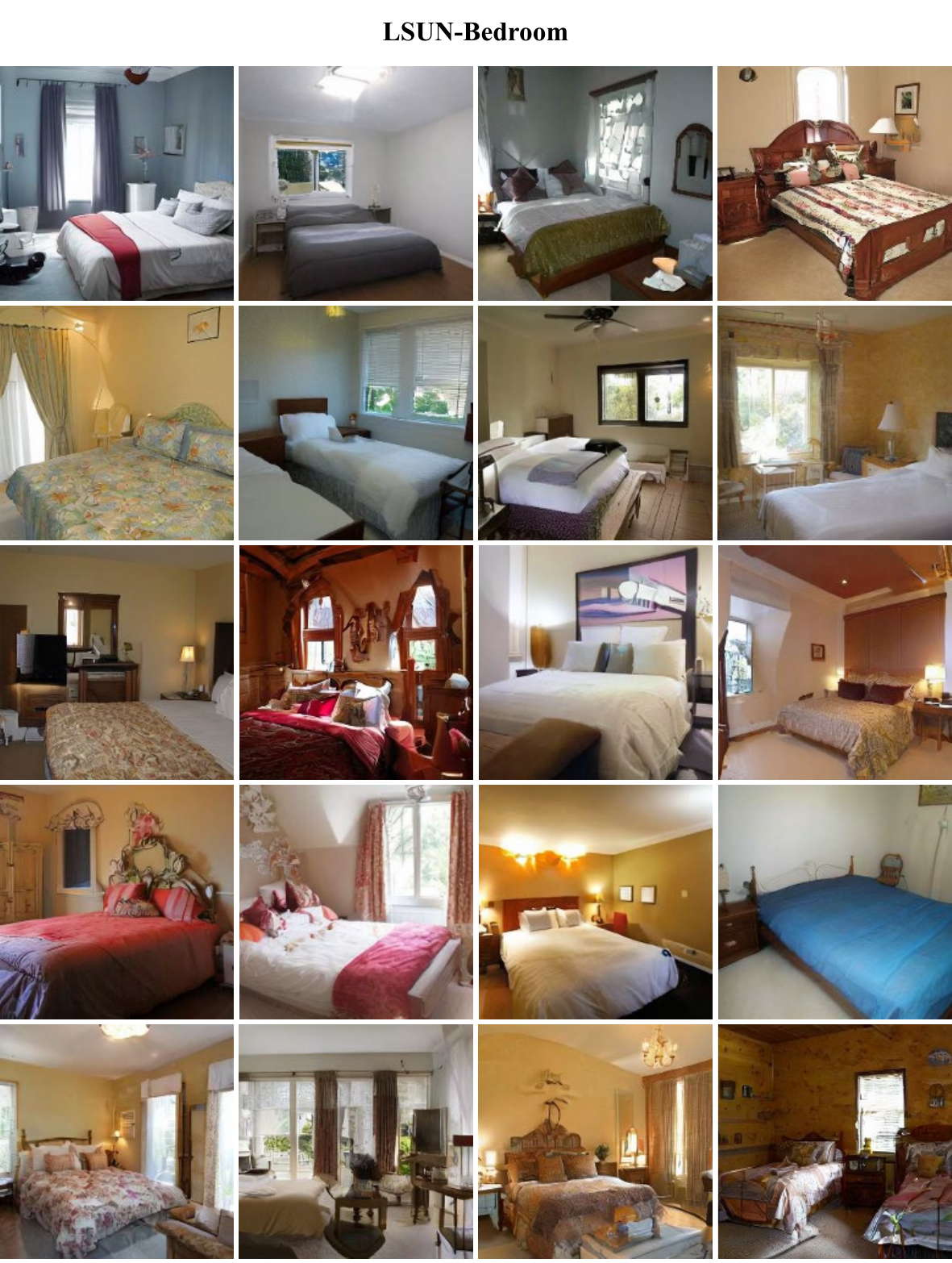}
  \caption{
 Visual results of image generation on the LSUN Bedroom datasets.
  }
  \label{fig:bedroom_gen_dirversity}
\end{figure*}

\begin{figure*}[t]
  \centering
  \includegraphics[width=0.9\linewidth]{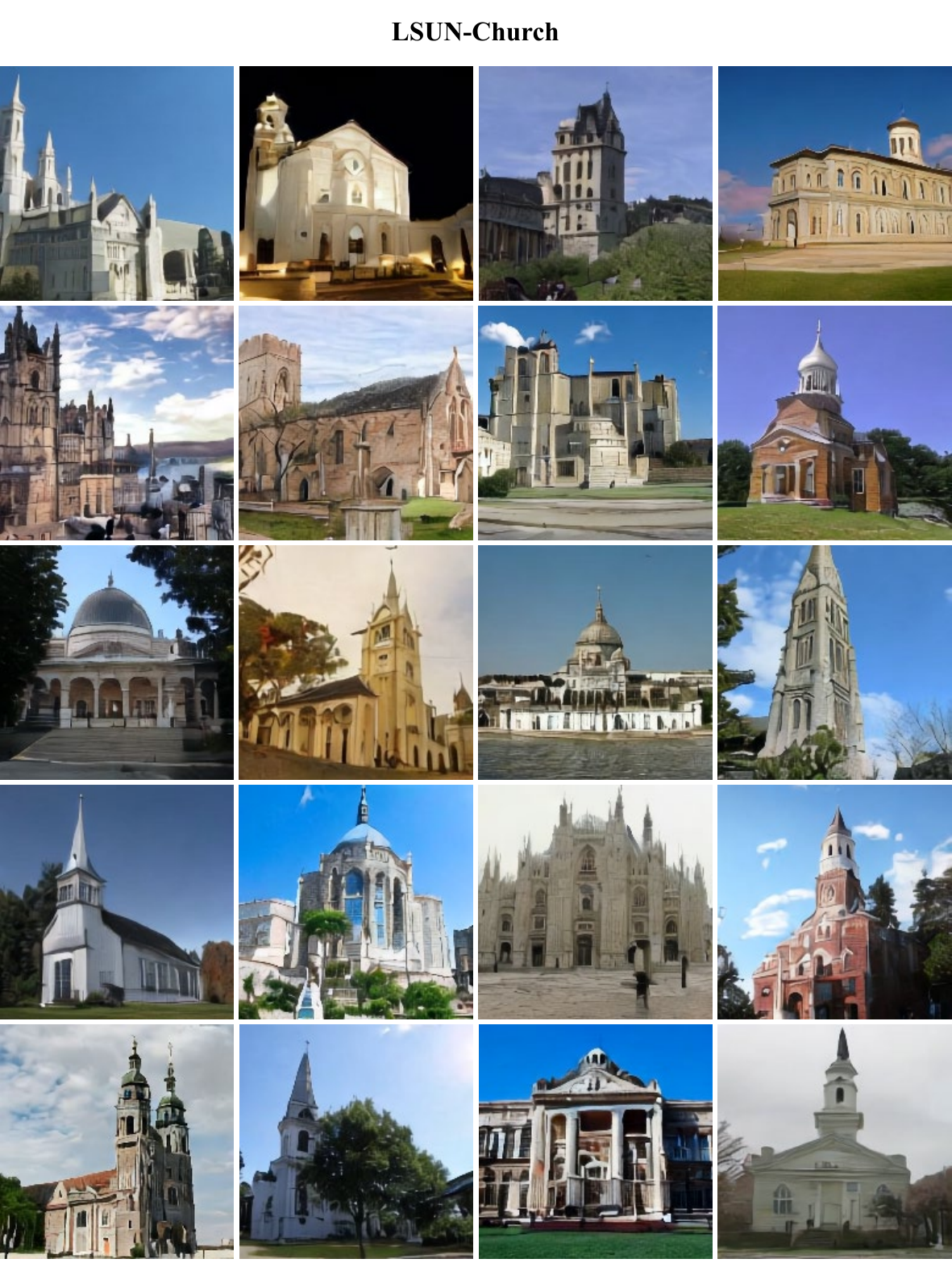}
  \caption{
  Visual results of image generation on the LSUN Church datasets.
  }
  \label{fig:church_gen_dirversity}
\end{figure*}

To demonstrate the performance of our model, we provide more provide more generated images and comparison results.
In \cref{fig:ffhq_gen_scale,fig:lsun_gen_scale,fig:ffhq_gen_dirversity,fig:bedroom_gen_dirversity,fig:church_gen_dirversity}, we visualize various randomly sampled results for the FFHQ, LSUN-Bedroom and LSUN-Church datasets, respectively.
Our model shows remarkable performance in synthesizing high-quality details with a variety of styles and scale-consistency.

\begin{figure*}[t]
  \centering
  \includegraphics[width=0.9\linewidth]{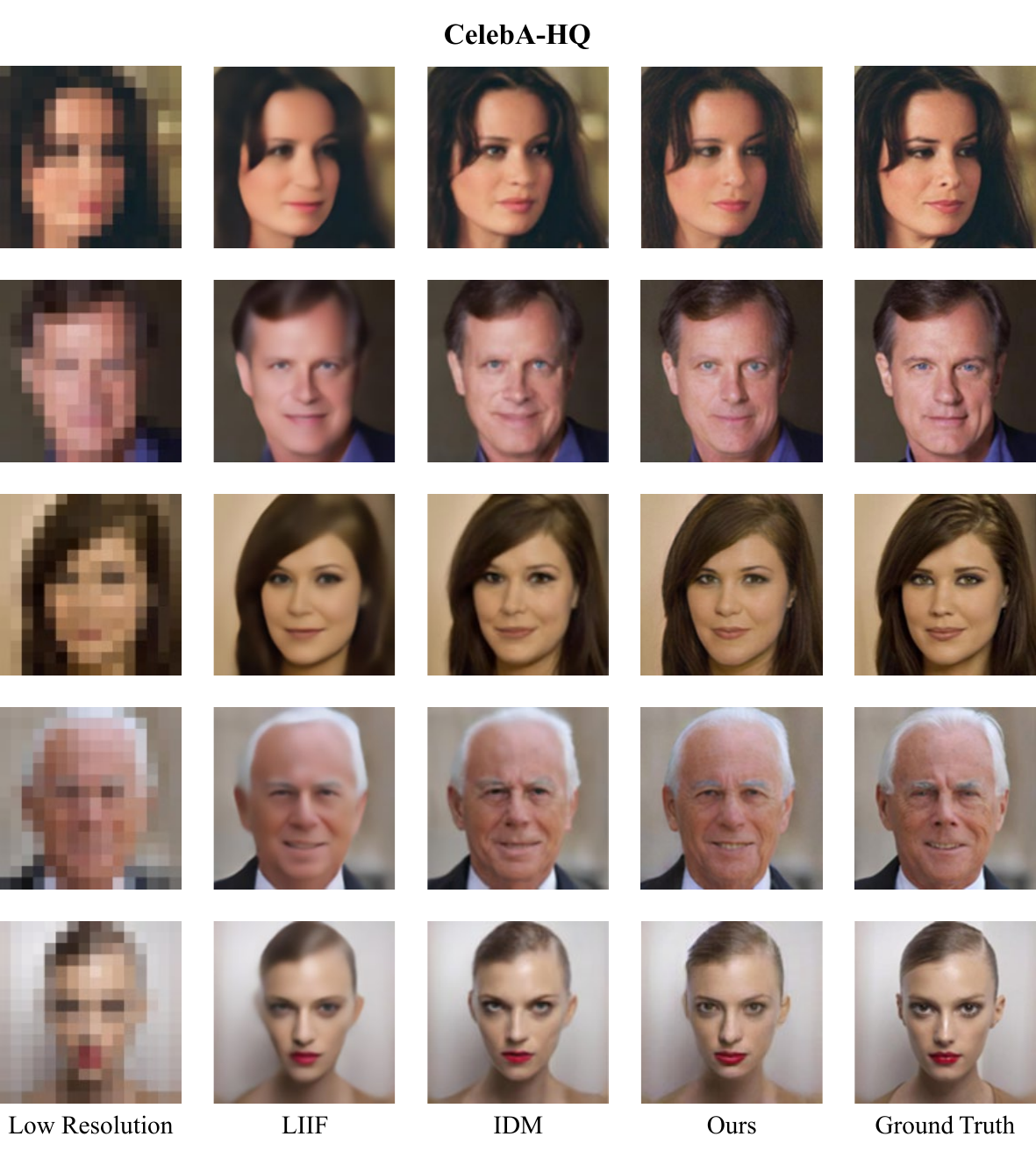}
  \caption{
  Comparison of $16\times16 \rightarrow 128\times128$ super-resolution on the CelebA-HQ datasets.
  }
  \label{fig:face_sr_comparison}
\end{figure*}

\begin{figure*}[t]
  \centering
  \includegraphics[width=0.9\linewidth]{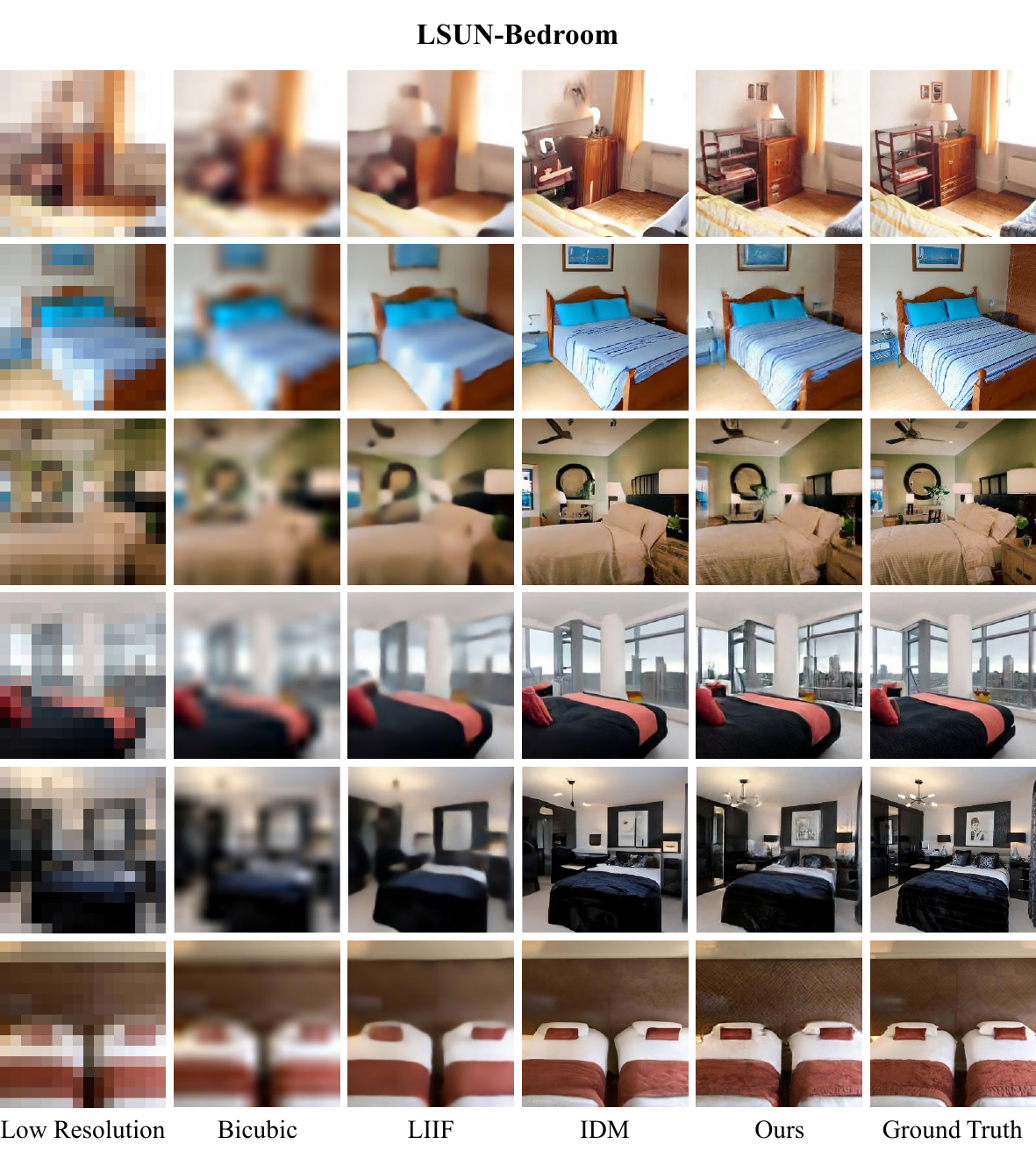}
  \caption{
  Comparison of $16\times16 \rightarrow 256\times256$ super-resolution on the LSUN Bedroom datasets.
  }
  \label{fig:bedroom_sr_comparison}
\end{figure*}

\begin{figure*}[t]
  \centering
  \includegraphics[width=0.9\linewidth]{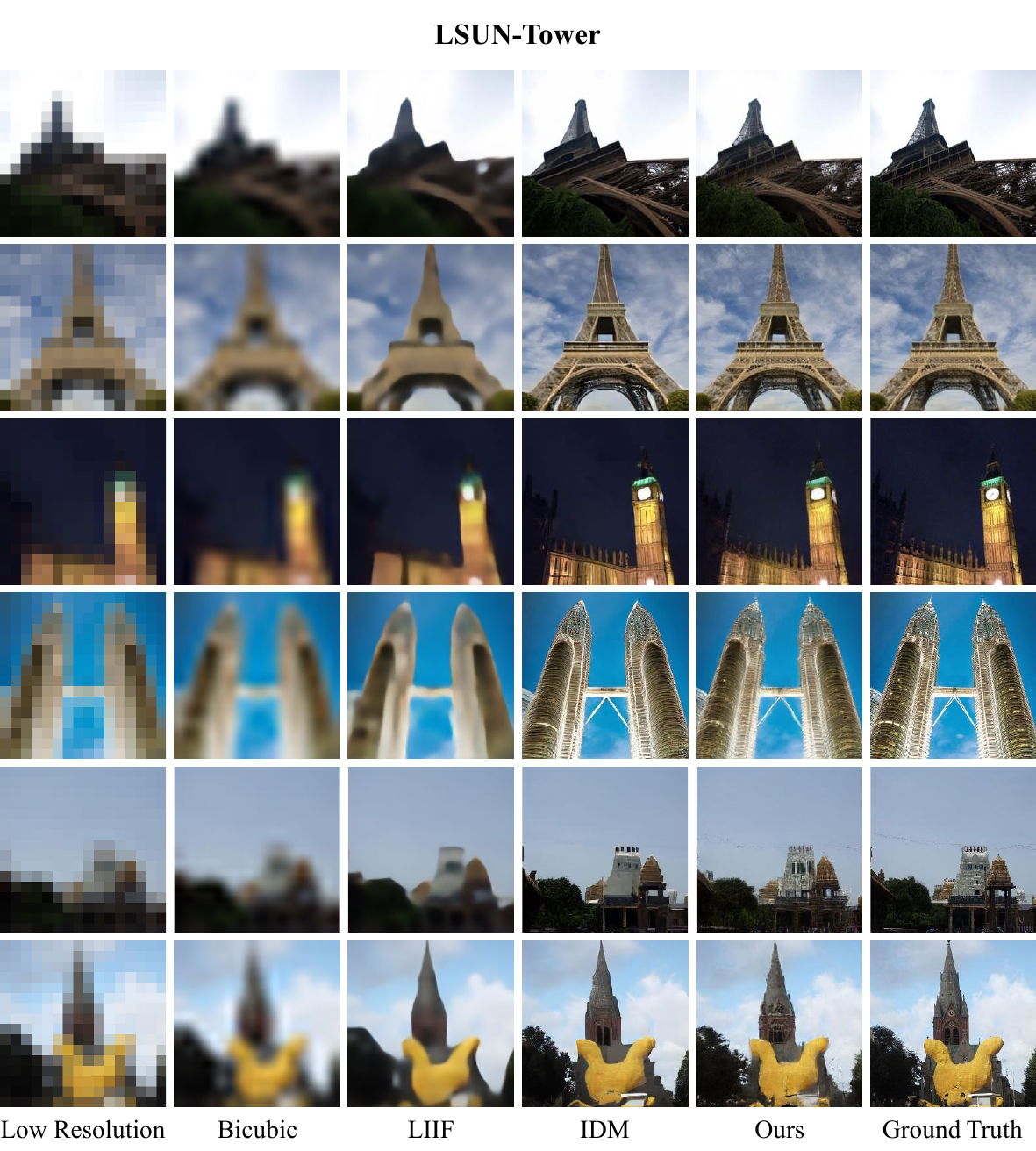}
  \caption{
  Comparison of $16\times16 \rightarrow 256\times256$ super-resolution on the LSUN Tower datasets.
  }
  \label{fig:tower_sr_comparison}
\end{figure*}

\begin{figure*}[t]
  \centering
  \includegraphics[width=0.9\linewidth]{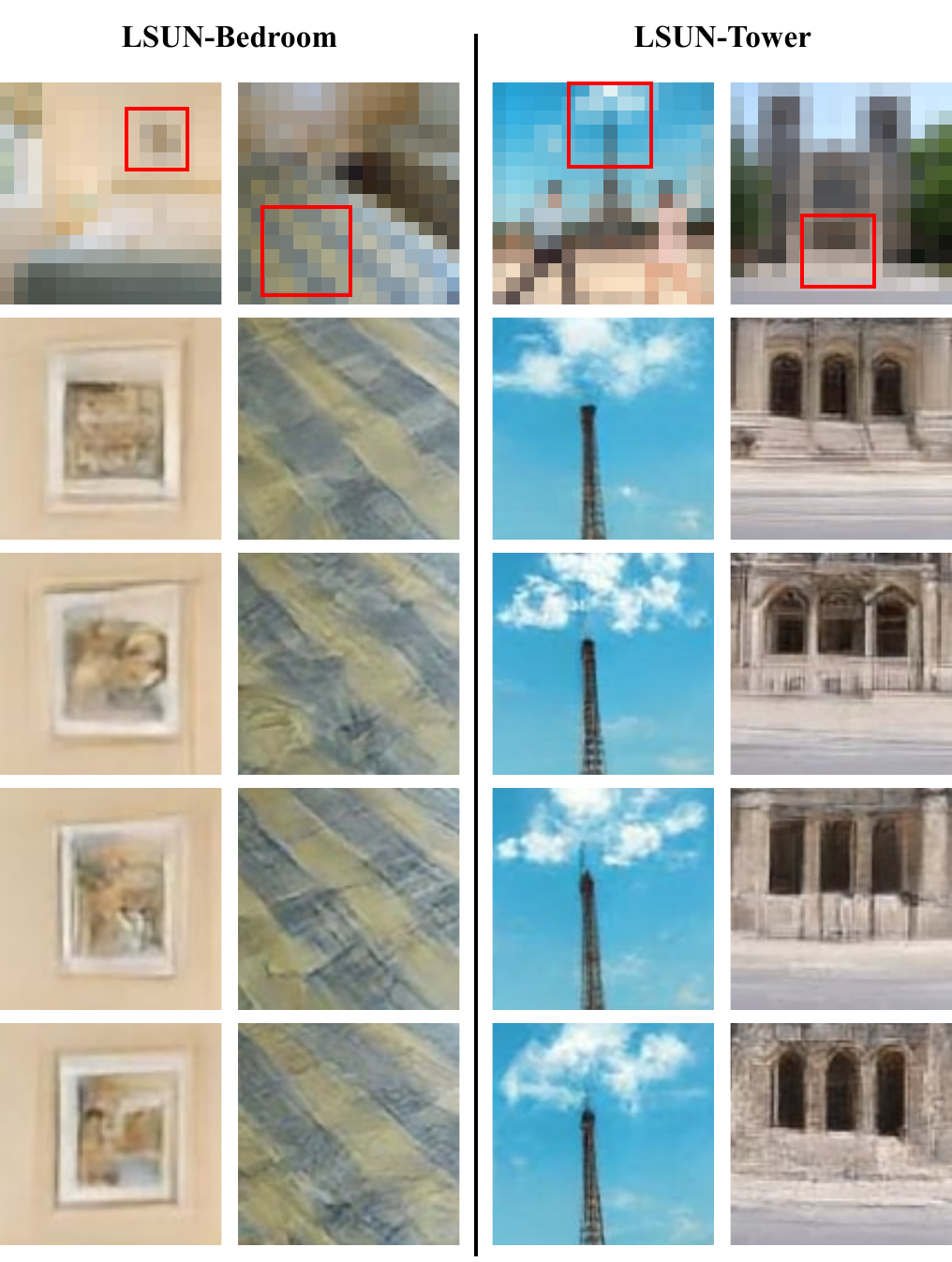}
  \caption{
  Visualization results of diversity in super-resolution tasks.
  The top image is an \textit{LR} image, and the images below are different \textit{SR} results in the red area.
  }
  \label{fig:sr_diversity}
\end{figure*}

\cref{fig:face_sr_comparison,fig:bedroom_sr_comparison,fig:tower_sr_comparison} show the qualitative comparison of \textit{SR} for CelebA-HQ~\cite{suppl_ProGAN}, LSUN-Bedroom and LSUN-Tower, respectively.
LIIF has over-smoothing issues in contrast to high PSNR scores.
Both IDM and our model are good at capturing high-resolution details, and furthermore, our model has achieved relatively few distortions.
In addition, \cref{fig:sr_diversity} shows various \textit{SR} results for the LSUN datasets.
The top image is an \textit{LR} image, and the images below are different \textit{SR} results in the red area.
As the scale increases, the number of high-resolution solutions that can be recovered from low-resolution becomes more diverse.
However, INR-based models such as LIIF~\cite{suppl_LIIF} always achieve only the same results.
In contrast, our stochastic model can generate a variety of patterns and textures for blankets, clouds, and buildings, etc. while maintaining \textit{LR} information.
This allows our model to better handle the `ill-posed problem'.

\end{document}